\documentclass[journal,twoside,web]{ieeecolor}
\usepackage{xcolor}
\usepackage{generic}
\usepackage[noadjust]{cite}
\usepackage{amsmath,amssymb,amsfonts}
\usepackage{graphicx}
\usepackage{algorithm}
\usepackage{textcomp}
\def\BibTeX{{\rm B\kern-.05em{\sc i\kern-.025em b}\kern-.08em
    T\kern-.1667em\lower.7ex\hbox{E}\kern-.125emX}}
\makeatletter
\let\NAT@parse\undefined
\makeatother
\usepackage[
    colorlinks=true,
    linkcolor=blue,
    citecolor=blue,
    urlcolor=blue
]{hyperref}

\usepackage{cleveref}

\usepackage{algpseudocode}

\usepackage{multirow, multicol}
\usepackage{nicematrix}
\usepackage{booktabs}
\usepackage{makecell}
\usepackage{color, colortbl}
\newcommand{\tcellbf}[1]{\textbf{\makecell{#1}}}

\usepackage{diagbox}

\usepackage{lmodern}

\usepackage[most]{tcolorbox}
\newtcolorbox{highlightbox}{
    colback=yellow!10!white,
    colframe=yellow!90!black,
    boxrule=0.5mm,
    sharp corners,
    boxsep=1pt,
    left=1pt,
    right=1pt,
    top=1pt,
    bottom=1pt,
    before upper={},
    after upper={},
    breakable
}

\newcommand{\track}[1]{\textcolor{blue}{#1}}
\newcommand{\revise}[2]{\textcolor{red}{\textbf{[#1]} #2}}

\usepackage{soul}

\renewcommand{\track}[1]{#1}
\renewcommand{\revise}[2]{#2}

\newcommand{\VarFirst}[0]{$\text{Var}_{1}$}
\newcommand{\VarRand}[0]{$\text{Var}_{\text{R}}$}

\begin{document}

\title{Adaptive Segmentation of EEG for\\Machine Learning Applications}

\author{Johnson Zhou, Joseph West, Krista A. Ehinger, Zhenming Ren, Sam E. John and David B. Grayden
\thanks{\textbf{\today}. This research was undertaken using the LIEF HPC-GPGPU Facility hosted at the University of Melbourne. This Facility was established with the assistance of LIEF Grant No. LE170100200. (Corresponding author: Johnson Zhou.)}
\thanks{Johnson Zhou, Joseph West and Krista A. Ehinger are with the School of Computing and Information Systems, The University of Melbourne, Parkville 3010 VIC, Australia (e-mail: johnson.zhou1@unimelb.edu.au, joseph.west@unimelb.edu.au, kris.ehinger@unimelb.edu.au)}
\thanks{Joseph West, Zhenming Ren, Sam E. John and David B. Grayden are with the Department of Biomedical Engineering, The University of Melbourne, Parkville 3010 VIC, Australia (e-mail: joseph.west@unimelb.edu.au, zhenmingr@student.unimelb.edu.au, sam.john@unimelb.edu.au, grayden@unimelb.edu.au).}
}

\maketitle

\begin{abstract}
\textit{Objective.}
Electroencephalography (EEG) data is derived by sampling continuous neurological time series signals. In order to prepare EEG signals for machine learning, the signal must be divided into manageable segments. The current naive approach uses arbitrary fixed time slices, which may have limited biological relevance because brain states are not confined to fixed intervals. We investigate whether adaptive segmentation methods are beneficial for machine learning EEG analysis.

\textit{Approach.}
We introduce a novel adaptive segmentation method, CTXSEG, that creates variable-length segments based on statistical differences in the EEG data and propose ways to use them with modern machine learning approaches that typically require fixed-length input. We assess CTXSEG using controllable synthetic data generated by our novel signal generator CTXGEN. While our CTXSEG method has general utility, we validate it on a real-world use case by applying it to an EEG seizure detection problem. We compare the performance of CTXSEG with fixed-length segmentation in the preprocessing step of a typical EEG machine learning pipeline for seizure detection.

\textit{Main results.}
We found that using CTXSEG to prepare EEG data improves seizure detection performance compared to fixed-length approaches when evaluated using a standardized framework, without modifying the machine learning method, and requires fewer segments.

\textit{Significance.}
\track{This work demonstrates that adaptive segmentation with CTXSEG can be readily applied to modern machine learning approaches, with potential to improve performance. It is a promising alternative to fixed-length segmentation for signal preprocessing and should be considered as part of the standard preprocessing repertoire in EEG machine learning applications.}
\end{abstract}


\section{Introduction}\label{sec:intro}

\IEEEPARstart{E}{lectroencephalography} (EEG) is a method to measure the electrical activity of the brain through electrodes that are positioned around or within the brain. EEG signals are time-series data that can be analyzed to extract features and patterns for identifying different states of the brain \cite{schomer_niedermeyers_2017}. In recent times, machine learning (ML) and deep learning (DL) algorithms have been increasingly used in the analysis of EEG \cite{Xi_2022_TwoStage,He_2022_Spatial,Thuwajit_2022_EEGWaveNet,srinivasan_detection_2023,qiu_lightseizurenet_2023,Zhu_2023_Automated,Song_2023_EEG,ingolfsson_minimizing_2024,awais_graphical_2024,li_end--end_2024,busia_reducing_2024,wang_epileptic_2024}.

To prepare EEG data for machine learning, continuous EEG is divided into discrete \textit{segments} in a preprocessing step known as \textit{segmentation} \cite{sanei_eeg_2007}. The current naive approach to segmentation is dividing the signal into fixed time slices, often arbitrarily; we refer to this approach as \textit{fixed-length segmentation}. However, this approach is likely to have limited biological relevance as brain states cannot be expected to occur at fixed intervals. Instead, we hypothesize that an approach that has segment boundaries relative to changes in the EEG is likely to be more informative for machine learning.

\textit{Adaptive segmentation} divides the EEG into \textit{variable-length} segments based on signal activity rather than fixed time intervals~\cite{sanei_eeg_2007}. This approach considers the concept of \textit{stationarity}, which refers to the statistical properties of a signal that remain relatively consistent over time, encouraging segments to align with potential neurological changes~\cite{eliasmith_neural_2004}. Although adaptive segmentation was conceptualized in the 1970s~\cite{remond_theorie_1972}, it has not been extensively used with modern machine learning approaches. We believe that this may be because variable-length segments may not be readily compatible with modern machine learning algorithms that typically require fixed-length inputs. To address this, \revise{R3-1}{we investigate the following Research Questions (RQ)}:

\begin{itemize}
    \item \track{\textbf{RQ1:}~How can the statistical properties of signal stationarity be used to extract (adaptive) variable-length EEG segments, that are statistically distinct in their signal activity, and create fixed-length representations of these segments to use with machine learning algorithms that expect fixed-length input?}
    \item \track{\textbf{RQ2:}~What is the efficacy of adaptive segmentation, relative to fixed-length segmentation, in terms of standardized performance benchmarks, when applied to EEG machine learning tasks?}
\end{itemize}

For \textbf{RQ1}, we propose a novel adaptive segmentation method called Context Segmentation (CTXSEG) that functions as part of the preprocessing step of an EEG machine learning pipeline and \revise{R1-1}{operates with linear complexity relative to signal length.} CTXSEG creates variable-length segments that are statistically distinct based on their frequency characteristics, a property that we refer to as \textit{context}. We leverage this contextual similarity to obtain a fixed-length representative window that is derived from the variable-length segments, making them compatible with machine learning (see \Cref{sec:method:segmentation}). 

To facilitate evaluation, we also develop a novel method to generate synthetic signals called Context Generation (CTXGEN) that is based on spiking neuron models. CTXGEN allows embedding user-definable changes to signal properties through a simple interface, while still capturing the random and noisy characteristics of real signals (see \Cref{sec:method:generation}). \revise{R1-4,R1-6}{We compare and contrast both CTXSEG and CTXGEN against existing methods in~\Cref{sec:exp:1}}, then seek to understand the segmentation characteristics of CTXSEG using CTXGEN in~\Cref{sec:exp:2}.

To assess the suitability and efficacy of CTXSEG for machine learning applications (\textbf{RQ2}), we replace fixed-length segmentation in the preprocessing step of a typical machine learning pipeline for seizure detection (see \Cref{sec:exp:3}). To our knowledge, this is the first application of adaptive segmentation \track{as a drop-in replacement to fixed-length segmentation in modern EEG machine learning approaches.} We demonstrate that CTXSEG can directly substitute fixed-length methods and find that \revise{R1-1,R2-1,E-2}{CTXSEG can improve the seizure detection performance of an existing machine learning algorithm when evaluated with a standardized evaluation framework}, especially when applied at test time. The key contributions of this work are:

\begin{itemize}
    \item Development of a new adaptive segmentation algorithm called Context Segmentation (CTXSEG) that is compatible with modern machine learning techniques,
    \item Development of a new, simple to use synthetic signal generator called Context Generation (CTXGEN), and
    \item Demonstration that using adaptive segmentation with CTXSEG can be used with modern machine learning approaches and a promising alternative to fixed-length segmentation for signal preprocessing. We recommend that such techniques be considered as part of the standard preprocessing repertoire in EEG machine learning applications.
\end{itemize}

The remainder of the paper is organized as follows. We begin with a brief background in \Cref{sec:background}, with a more detailed review of existing adaptive segmentation methods in~\Cref{sec:background:adaptive_segmentation}. We present our proposed methods for CTXSEG and CTXGEN in \Cref{sec:methods}. Experimental results are detailed in \Cref{sec:results} followed by discussion in \Cref{sec:discussion} and concluding remarks in \Cref{sec:conclusion}. Source code and installable Python package for both CTXSEG and CTXGEN can be found online at: \hyperlink{https://github.com/johnsonjzhou/ctxseg}{https://github.com/johnsonjzhou/ctxseg}.

\section{Background}\label{sec:background}


\textbf{EEG time-series signals:} EEG signals are a record of the electrical activity of the brain over time. When analyzing EEG, they are often considered as sampled time series $\mathbf{x} = \left\{x_{n} \right\} : n = 0,1, ..., N-1$; i.e.,  a set of samples $x_n$ over the total recording duration $T$. The duration $T$ is related to the number of samples $N$ and the sampling interval $\Delta t$ by $T=N \cdot \Delta t$. Here, $n$ is an integer representing the sample index and the sampling interval $\Delta t$ is related to the sampling frequency $f_s$ (Hz) by the equation $\Delta t = {1}/{f_s}$ \cite{brockwell_time_1991,oppenheim_discrete-time_1999}.

\textbf{EEG-signals are non-stationary:}  EEG signals are \textit{non-stationary} as the fluctuations observed in the signals are highly random and can change over time \cite{kosar_classification_2004,schomer_niedermeyers_2017}. Stationarity in signals refers to whether the statistical properties of the samples within time series signals remain unchanged over time.
Formally, a time series is stationary if the mean $\mu$ is constant, such that $\mathbb{E}(x_t) = \mu$ for all $t \in T$, and the covariance between two samples depends only on the time difference (or \textit{lag} denoted as $h$) and not on the specific time $t$, such that $\gamma(h) = \text{Cov}(x_{t+h}, x_t)$, where $\gamma(\cdot)$ is an auto-covariance function \cite{brockwell_time_1991}. This type of stationarity is also referred to as wide-sense stationarity.

\textbf{Segments, windows and epochs:} EEG signals can be continuously measured over sampling periods ranging from a few seconds to as long as days or even years. In order to analyze long-duration continuous EEG, it must be divided into manageable chunks in a preprocessing step known as \textit{segmentation} \cite{sanei_eeg_2007}. Some EEG machine learning works use \textit{window} and \textit{segment} terminology interchangeably. Adaptive segmentation methods may involve multiple stages of combining EEG chunks. For disambiguation, we define a \textit{window} as a sequence of samples with a fixed length and a \textit{segment} as the final combined sequence of windows that can be fixed or variable in length. Note that our use of the term \textit{window} differs from that of a \textit{window function}, such as Hamming window that is used when performing spectral analysis \cite{harris_windows_1978}. 

Our definition of the term \textit{segment} implies that EEG segments are used as inputs to computational algorithms for analysis. These segments are sometimes referred to as \textit{epochs} in the context of EEG processing \cite{sanei_eeg_2007}, which conflicts with the machine learning definition of training epochs. We maintain a distinction in terminology to avoid confusion with training epochs in the context of machine learning, which refers to a complete pass through the training dataset.

\textbf{Fixed segmentation for machine learning:} Machine learning, including deep learning, is gaining popularity for many EEG analysis tasks such as seizure detection (\Cref{tab:recent_studies}). Most machine learning EEG approaches use fixed-length segmentation, with the window size and stride being a point of study, and either empirically or arbitrarily selected.    

\textbf{Adaptive segmentation:} The concept of adaptive segmentation was first introduced in the 1970s with the theory of graphic objects (also known as grapho-elements) of EEG signals \cite{gersch_spectral_1970,remond_theorie_1972} \revise{R3-7}{with applications in extracting and analyzing elementary patterns of EEG~\cite{bodenstein_feature_1977,Praetorius_1977_Adaptive,Barlow_1981_Automatic,Jansen_1981_Piecewise,appel_adaptive_1983}, analysis of sleep patterns~\cite{gath_computerized_1980,gath_classical_1983,Barlow_1981_Automatic,Putilov_2007_Segmental}, visual evoked responses~\cite{Gath_1985_Automatic}, as well as epileptic seizures~\cite{varri_computerized_1988,Pietila_1994_Evaluation,wendling_segmentation_1997,kosar_classification_2004}.} These are elementary patterns within the signals where the samples in each segment are \textit{quasi-stationary}, meaning the spectral properties of the samples are nearly stationary for a short duration of time \cite{bodenstein_feature_1977}. The boundary between two different quasi-stationary segments is often referred to as the \textit{segment boundary} \cite{sanei_eeg_2007}. In the context of adaptive segmentation, (input) segments refer to the quasi-stationary segments that may vary in size, and windows refer to subsegments that are used to discover segment boundaries. We equate stationarity with EEG context within the our CTXSEG approach. CTXSEG achieves contextual segmentation by seeking to ensure that the segment is quasi-stationary. A review of existing adaptive segmentation methods is presented in~\Cref{sec:background:adaptive_segmentation}.

\textbf{Synthetic test signals:} Adaptive segmentation methods are typically evaluated using synthetic signals rather than real EEG. This is because obtaining ground-truth labels for segment boundaries is often a challenge in real signals \cite{appel_comparative_1984}. \textit{Ground-truth} labels refer to segment boundaries for which a change in signal property is known. For example, real EEG is often annotated for changes in seizure states, but rarely also labeled for artifacts from muscle and eye movement or electrical interference from recording equipment and other electrical sources. This limitation necessitates the use of synthetic test signals with known labels for method verification. Synthetic signals are commonly generated by combining multiple harmonics~\cite{Azami_2012_New,Azami_2012_Improved,Azami_2013_hybrid,Azami_2014_New,Azami_2014_Adaptive,Azami_2015_intelligent} or auto-regressive models that are fitted to real EEG~\cite{appel_comparative_1984,aufrichtigl_adaptive_1991,wendling_segmentation_1997} (see~\Cref{appendix:exp1} for more details).

\section{Methods}\label{sec:methods}

\subsection{Method of adaptive segmentation}\label{sec:method:segmentation}

\begin{figure*}
    \centering
    \includegraphics[width=1\linewidth]{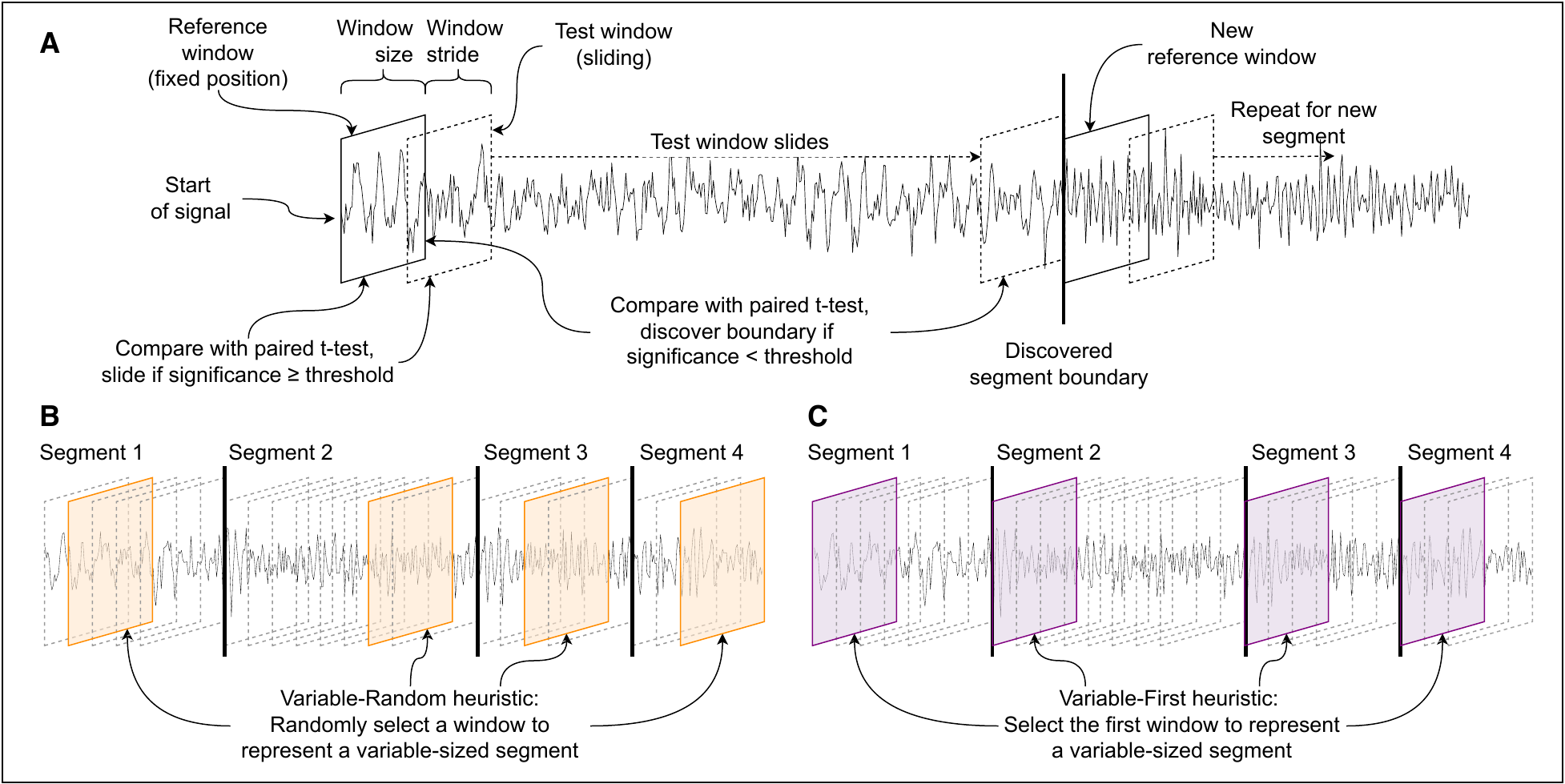}
    \caption{
    Schematics showing the operation of the proposed Context Segmentation (CTXSEG) segmentation method (see \Cref{sec:method:segmentation}). 
    \textbf{A:} Process of adaptive segmentation in CTXSEG. 
    The frequency spectrum of a stationary reference window (solid line) is compared with the frequency spectrum of a sliding test window (dotted line) using the paired t-test. A segment boundary is discovered if the level of significance is below a predefined threshold, indicating that the frequency spectrum between the reference and test windows are significantly different. Once a segment boundary has been discovered, the reference window advances to the first window of the new segment and the procedure is repeated. This creates variable-size segments where all windows within the segment are statistically similar in the frequency domain.
    \textbf{B,C:} Two heuristics used to select a fixed-size window to represent a variable-sized segment, where each fixed-sized window is statistically similar in the frequency domain (see~\Cref{sec:method:segmentation} and \Cref{tab:ctxseg:strategies}). Variable-Random (B), a random window is selected from each segment. For Variable-First (C), the first window is selected from each segment.
    }
    \label{fig:seg}
\end{figure*}

We introduce Context Segmentation (CTXSEG), a novel non-parametric method for adaptive segmentation that uses a statistical distance measure across two sliding windows. CTXSEG has three user-definable parameters, window size ($w$), stride ($s$) and significance threshold ($\alpha$), and runs with linear complexity relative to the size of the signal (see~\Cref{alg:ctxseg}). \revise{R1-4}{In contrast to existing approaches~\cite{varri_computerized_1988,agarwal_adaptive_1999,Jakaite_2011_Feature}, we statistically compare the full frequency spectra rather than select frequency bands to maintain general applicability. In addition, we fix the reference window and only slide the test window to reduce discovery of redundant segment boundaries, a well-known shortcoming of contiguous windows approaches~\cite{krajca_automatic_1991}.} 

Since the signal properties within the discovered segments are statistically similar, we refer to them as \textit{contextual segments}. \revise{R3-3}{We consider these segments to be quasi-stationary due to their statistical similarity~\cite{bodenstein_feature_1977,Cestari_2017_Stochastic}.} There are further research opportunities here in developing new methods to refine the identification of segment boundaries, which may improve the effectiveness of quasi-stationarity for various applications.

\begin{algorithm}[t]
    \caption{Segmentation by Context (CTXSEG)}  
    \label{alg:ctxseg}
    \begin{algorithmic}
        \Statex \textbf{Input:} $\mathbf{x} \in \mathbb{R}^N$: EEG-signal of size $N$,
        \Statex \quad $w$: Window size,
        \Statex \quad $s$: Stride,
        \Statex \quad $\alpha$: Significance threshold.
        \Statex \textbf{Output:} $\mathbf{z} \in \{\text{False}, \text{True}\}^N$: Segment boundaries of size $N$ (True for boundary positions, False otherwise).
    \end{algorithmic}
    \begin{algorithmic}[1]
    \Procedure{CTXSEG}{}
        \State Initialize segment boundaries $\mathbf{z} \gets \textbf{False}^{N}$
        \State Reference pointer $r \gets 0$
        \State Test pointer $t \gets s$
        \While{$t < (N-w)$}
            \State Reference window $\mathbf{w_r} \gets \mathbf{x}[r, r+w]$
            \State Test window $\mathbf{w_t} \gets \mathbf{x}[t, t+w]$
            \State Apply window function to $\mathbf{w_r}$ and $\mathbf{w_t}$
            \State $p \gets$ \Call{t-test}{$\log(\mathcal{F}(\mathbf{w_r}))$, $\log(\mathcal{F}(\mathbf{w_t}))$}
            \If{$p < \alpha$}
                \Comment{Boundary discovered}
                \State Boundary position at $b \gets t + w$
                \State Assign boundary $\mathbf{z}[b] \gets \text{True}$
                \State Advance $\mathbf{w_r}$: $r \gets b + 1$
                \State Advance $\mathbf{w_t}$: $t \gets r + s$
            \Else
                \State Slide $\mathbf{w_t}$: $t \gets t + 1$
            \EndIf
        \EndWhile \\
        \Return Segment boundaries $\mathbf{z}$
    \EndProcedure
\end{algorithmic}
\end{algorithm}

\textbf{Statistical distance measure:} We use a statistical test as the non-parametric distance measure to compare the frequency spectra of the \textit{reference} ($\mathbf{w_r}$) and \textit{test} ($\mathbf{w_t}$) windows (see \Cref{alg:ctxseg} and \Cref{fig:seg}). We primarily use the paired t-test because of its simplicity and comparable performance to other statistical tests in our initial evaluations. However, we acknowledge that more specialized tests may be more suitable for specific applications~\cite{Jakaite_2011_Feature}.


\track{The frequency content of the two windows can be considered statistically different, therefore indicative of a segment boundary, if the p-value falls below a significance threshold that is defined by the user as a parameter ($\alpha$). In our experiments, we consider $\alpha=$ 0.05, 0.01 and 0.001, representing common p-value significance levels for statistical testing in the medical domain~\cite{Chu_1999_introduction,Cesana_2018_What,Ioannidis_2018_Proposal}.}

We compute the frequency spectra of each window by first applying an appropriate windowing function, such as the Hamming window, and transforming into the frequency domain using the Fast Fourier Transform (FFT). We use the log magnitude of the resulting spectra for comparison by the paired t-test to help balance the representation of higher frequencies, which typically have lower amplitudes.

\textbf{Sliding-window operation:} The sliding-window algorithm for the test window operates similarly to the Spectral Error Measure (SEM) method \cite{bodenstein_feature_1977} and is schematically represented in \Cref{fig:seg}. A user can define the parameters of window size $w$, which determines the sizes of both the reference and test windows, and stride $s$, which is the step size at which the test window moves. At each step, the log frequency spectra of the windows are compared using the paired t-test. A segment boundary is detected at the position of the test window when the reference and test windows are found to be statistically different. 

When a segment boundary is detected, the reference window is advanced to the next non-overlapping position (with the leading test window) to ensure that signal properties are independent between segments. Due to this, the minimum segment size equals the window size. The overall time complexity is linear with respect to the number of windows, and both the reference and test windows remain equal in size throughout the entire operation. 

Keeping the reference window fixed while the test window slides is shown to be more robust to slow changes in signal property (trends) when compared with methods that use two contiguous windows \cite{krajca_automatic_1991}. The two contiguous windows method was initially considered due to an apparent difficulty in keeping windows synchronized when simultaneously segmenting across multi-channel EEG \cite{krajca_online_1991}. While this may have been a limitation at the time (circa 1988), window synchronization can easily be achieved in most modern programming languages. 


\textbf{Using variable-length segments for machine learning:} Contextual segments created using CTXSEG have the following three characteristics: 

\begin{enumerate}
    \item every window within each contextual segment are statistically similar in the frequency domain up to a predefined level of significance, 
    \item contextual segments vary in size and are at a minimum the window size, and
    \item contextual segments do not overlap with adjacent segments.
\end{enumerate}

Variable-length segments raises a challenge for machine learning methods, as most require fixed-length inputs. In order to address this challenge, we ask the question: Is there a fixed size window that effectively represents the entire segment? \revise{R2-6,R3-4}{We propose that since the entire segment has similar context, this property may be suitable to serve as a heuristic to extract fixed-length representations of variable-length segments. Specifically, we consider selecting the first window (referred to as \textit{Variable-First}, denoted \VarFirst, see~\Cref{fig:seg}C) or a random window (referred to as \textit{Variable-Random}, denoted \VarRand, see~\Cref{fig:seg}B) to represent the entire variable-length segment. There are further research opportunities here in finding the best fixed-length representation of these variable-length segments.}

In the following, \revise{R2-5}{we propose five strategies for using adaptive segments with machine learning (see~\Cref{tab:ctxseg:strategies}). Strategies \VarFirst-\VarFirst\ and \VarRand-\VarFirst\ use CTXSEG for both model training and inference. For inference, we consistently select the first window of each segment (i.e., Variable-First) to allow for earliest possible presentation of each variable-length segment. It is also the only practical solution if CTXSEG is used in online settings since the end of the segment may not be known in advance. For strategies \VarFirst-Fixed and \VarRand-Fixed, we use CTXSEG for training but not inference. For the strategy Fixed-\VarFirst, we retain existing (fixed) segmentation methods for training and only use CTXSEG for inference.} \revise{R2-6}{For the absence of doubt, our consideration of signal context is only applicable for signal segmentation in preprocessing and defer matters of feature extraction and/or classification to machine learning methods downstream.} 

\begin{table}[h]
    \caption{Strategies for using variable-length segments\\with machine learning}
    \centering
    \scalebox{1.0}{
    \begin{tabular}{ccc}
        \hline
        \textbf{Strategy}&
        \textbf{Training}&
        \textbf{Inference}\\

        \hline
        Fixed-Fixed (baseline)&
        Fixed&
        Fixed\\
        
        \hline
        \VarFirst-\VarFirst&
        Variable-First&
        Variable-First\\

        \hline
        \VarRand-\VarFirst &
        Variable-Random&
        Variable-First\\
        
        \hline
        \VarFirst-Fixed&
        Variable-First&
        Fixed\\
        
        \hline
        \VarRand-Fixed&
        Variable-Random&
        Fixed\\

        \hline
        Fixed-\VarFirst&
        Fixed&
        Variable-First\\

        \hline
        
    \end{tabular}
    }
    \label{tab:ctxseg:strategies}
\end{table}

We investigate the feasibility of our proposed strategies in an EEG machine learning-based seizure detection task in~\Cref{sec:exp:3} and find that CTXSEG can be used as a drop-in replacement to existing fixed size segmentation to improve the performance of existing machine learning methods in certain circumstances.

\subsection{Method of generating signals}\label{sec:method:generation}

We introduce a novel method for generating synthetic test signals that we refer to as Context Generation (CTXGEN), \track{which generates stochastic signals based on user defined neuronal firing rates $f_r$ at any given time. Users can define \textit{context states}, representing ground-truth segment boundaries, by maintaining a constant firing rate for a fixed duration. For the purposes of assessing suitability of contextual segmentation approaches, these synthetic context states are analogous to brain states that exhibit stable neural activity patterns over short periods. A segment containing one context state is quasi-stationary and changing the firing rate introduces non-stationarity to the signal.}

\begin{figure}[t]
    \centering
    \includegraphics[width=1.0\linewidth]{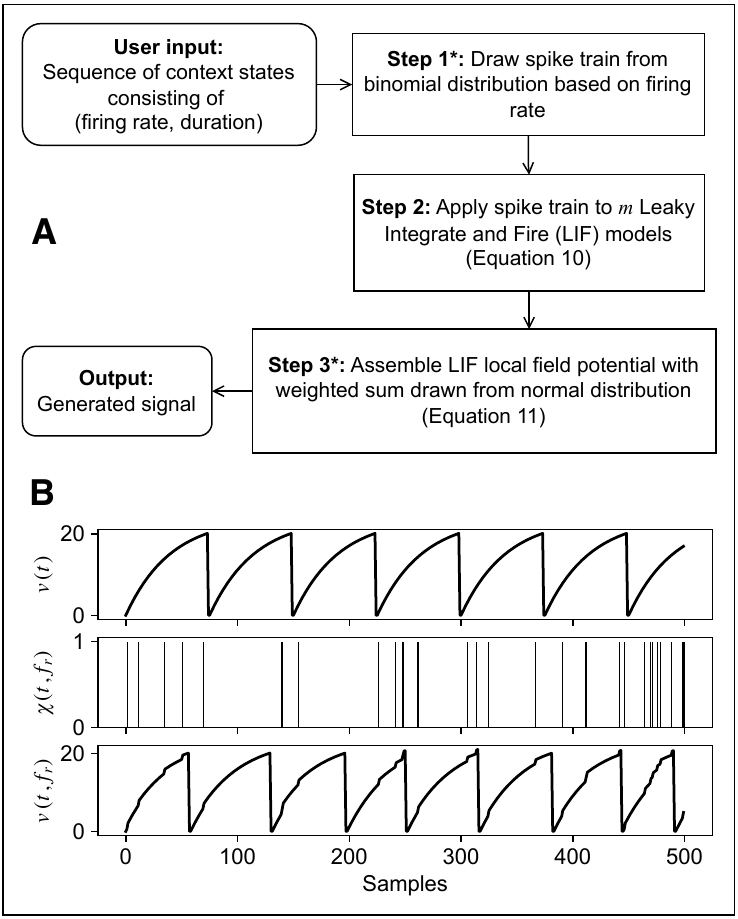}
    \caption{\textbf{A:} \revise{R1-5}{Schematic for the process of generating signals with Context Generation (CTXGEN). Noise is added at Steps 1 and 3.} 
    \textbf{B:} Graphical representation of neuronal local field potential (LFP) in CTXGEN when generating signals with $v_{\text{thresh}}=20$. (upper) LFP of the standard leaky integrate-and-fire (LIF) model. (middle) Spike train drawn from binomial distribution with probability dependent on firing rate. (lower) LFP that combines the standard LIF model with the binomial spike train so that neural spikes can be simulated to be dependent on firing rate.}
    \label{fig:ctxgen}
\end{figure}

The method uses multiple spiking neuron models to simulate local field potentials (LFP) based on user-defined neuronal firing rates. Simulated test signals are assembled by combining the LFPs of $m$ models and adding noise (see~\Cref{fig:ctxgen}A). We use $m=500$, although this hyperparameter can be defined by the user. This method improves upon existing approaches as it allows users to gradually vary the signal context by adjusting only the firing rate, avoiding the need for multiple auto-regressive models, while the added noise enhances the realism of the signals. 

Spiking neuron models are a family of conductance-based models that are highly representative of neuronal electrophysiology \cite{gerstner_spiking_2002}. The leaky integrate-and-fire (LIF) model is a well known spiking neuron model in which membrane potential $v$ increases over time from an initial state $v(t) = 0$ to a spiking state $v(t) = v_{\text{thresh}}$ at which a spike is fired at the time when the threshold voltage $v_{\text{thresh}}$ is reached, and the potential is reset to the initial state. The dynamics of the membrane potentials in LIF models are generally described by
\begin{equation}\label{eq:lif}
    v(t) = I(t)R - \tau \frac{\text{d}v}{\text{d}t},    
\end{equation}
where $I(t)$ is the injection current at time $t$, $R$ is the resistance and \track{$\tau$ is a constant that governs how quickly the membrane potential changes over time~\cite{stein_models_1967}.}


We assume that intrinsic neuron stimulation is constant, while extrinsic stimulation from the spiking activity of nearby connected neurons can have an additive effect. The dynamics of randomly connected spiking neurons can generally be approximated by a binomial distribution \cite{roxin_role_2011, rijal_exact_2023}. We simulate the random neuronal firing by drawing the spike train $\chi(t, f_r)$ from a binomial distribution with probability based on the firing rate of the neuron, such that $\chi(t, f_r) \sim (1, f_r \cdot \Delta t)$.

Neuronal activity is simulated by combining the LIF model with a binomial spike train. The local field potentials of individual neurons $v(t, f_r)$ are calculated for each time step a
\begin{equation}\label{eq:membrane_eq}
    v(t, f_r) = 
        \begin{cases}
            v(t) + \chi(t, f_r), & \text{if } v(t, f_r) < v_{\text{thresh}} \\
            0, & \text{otherwise}.
        \end{cases}
\end{equation}

We approximate and assemble the generated signal $x(t, f_r)$ by taking a weighted sum of the local field potentials from $m$ simulated neurons \cite{mazzoni_computing_2015}. With $\mathbf{V}$ representing $m$ local field potentials and $\mathbf{W}$ weights of size $m$ drawn from a normal distribution $\mathcal{N}(0,1)$, the generated signal can be constructed by
\begin{equation}
    x(t, f_r) = \mathbf{W}^{\mathsf{T}} \cdot \mathbf{V}, \quad \mathbf{W} \sim \mathcal{N}(0,1).
\end{equation}

Our proposed method is therefore able to generate signals based on user-defined neuronal firing rate $f_r$ as the source for contextual variation. The inclusion of noise via the two random elements, the spike-train and weighted sum in signal assembly, allows the signals to be more representative of the variability and unpredictability seen in real signals.

\subsection{Evaluation metrics}\label{sec:method:eval_simulated}

We consider the measures of boundary delay, boundary sensitivity and boundary similarity for evaluating with synthetic signals and \revise{R2-4}{use the standardized Seizure Community Open-Source Research Evaluation (SzCORE) framework~\cite{Dan_2024_SzCORE} to evaluate seizure detection performance.}

\textbf{Boundary delay:} The criterion for boundary delay is the time difference, in seconds, between a true segment boundary and a subsequently discovered segment boundary. This is also referred to simply as \textit{delay} in some studies \cite{appel_comparative_1984,krajca_online_1991}. Boundary delay is, in principle, the same as detection latency for seizure detection, where the true segment boundary is representative of the time when the brain state changes from non-seizure to seizure~\cite{bruno_seizure_2020}. This metric is important for adaptive segmentation as it represents the earliest opportunity for a seizure detection algorithm to detect a seizure from a segment of EEG. Boundary delay is not meaningful for fixed-length segmentation, because it is not dependent on the properties of the signal. 

\textbf{Boundary sensitivity:} It is possible that boundary delay can be excessively large such that it exceeds the amount of time that a signal remains stationary. For example, suppose that the duration between two ground-truth segment boundaries A and B is 1 second, but the boundary delay for boundary A is 1.5 seconds. In this situation, we consider that boundary discovery for A has been missed and the boundary delay for B is 0.5 seconds. Missed boundary discovery in adaptive segmentation has not been quantitatively evaluated in prior studies. To address this, we propose a new metric, referred to as \textit{boundary sensitivity}, that measures the proportion of discovered segment boundaries that were not missed for a given ground truth boundary.

\textbf{Boundary similarity}, proposed by~\cite{Fournier_2013_Evaluating}, measures the degree of agreement between ground truth and discovered segment boundaries on a scale of 0 to 1, where 1 signifies the discovered boundaries are identical to the ground truth and 0 being completely different. It takes into account the edit distance~\cite{Fournier_2012_Segmentation}, which applies a penalty for both under- and over-segmentation, as well as accounting for near misses within an acceptable tolerance. Although originally developed for text, it has been used in other domains such as audio processing~\cite{Wang_2021_Soloist,Ghinassi_2023_Exploring} and automated speech analysis~\cite{Moore_2016_Automated}.

\revise{E-1,R2-1}{\textbf{SzCORE framework}, proposed by~\cite{Dan_2024_SzCORE}, is a standardized benchmark for evaluating seizure detections that supports several commonly studied datasets. Under this framework, performance is assessed by conducting sample- and event-based scoring in both subject-specific (i.e., personalized) and subject-independent scenarios, with the objective of maximizing clinical relevance.}

\section{Results}\label{sec:results}

We evaluate the efficacy of our proposed adaptive Context Segmentation (CTXSEG) method using synthetic signals (Sections \ref{sec:exp:1} and \ref{sec:exp:2}) as well as real EEG data in a machine learning-based seizure detection task (\Cref{sec:exp:3}).

\begin{table*}[!ht]
\caption{Comparison of adaptive segmentation using synthetic signals}
\centering
\scalebox{0.90}{
\begin{tabular}{l|rrrr|rrrr|rrrr}
\toprule
  
  \multirow{2}{*}{\diagbox{Method}{Signal}} & \multicolumn{4}{c|}{Harmonics} & \multicolumn{4}{c|}{Auto-regressive} & \multicolumn{4}{c}{\textbf{(Ours) CTXGEN}} \\
  &   Count(6) &   Delay$\downarrow$ &   Sens$\uparrow$ &   Sim$\uparrow$ &  Count(6) &  Delay$\downarrow$ &   Sens$\uparrow$ &   Sim$\uparrow$ &   Count(6) &   Delay$\downarrow$ &   Sens$\uparrow$ &   Sim$\uparrow$ \\
              
\midrule
  Varri &            
  244.00 &              
  0.06 &           
  \textbf{1.0000} &          
  0.0240 &     
  773.73 &       
  \textbf{0.03} &    
  \textbf{1.0000} &   
  0.0077 &       
  1,113.81 &           
  \textbf{0.02} &        
  \textbf{1.0000} &       
  0.0053 \\
  
\midrule  
 NLEO &            
 106.00 &              
 0.11 &           
 \textbf{1.0000} &          
 0.0562 &     
 694.84 &       
 0.04 &    
 \textbf{1.0000} &   
 0.0087 &       
 1,207.16 &           
 \textbf{0.02} &        
 \textbf{1.0000} &       
 0.0049 \\

\midrule
 SPS 0.05 &          
 1,594.00 &              
 \textbf{0.04} &           
 \textbf{1.0000} &          
 0.0036 &     
 278.52 &       
 1.94 &    
 0.8650 &   
 0.0176 &          
 95.54 &           
 4.40 &        
 0.6025 &       
 0.0194 \\
 
 SPS 0.01 &             
 \textbf{56.00} &              
 4.28 &           
 0.5000 &          
 0.0442 &      
 60.54 &       
 8.02 &    
 0.4097 &   
 0.0296 &           
 7.92 &          
 14.65 &        
 0.0792 &       
 0.0163 \\
 
 SPS 0.001 &              
 0.00 &             
 17.50 &           
 0.0000 &          
 0.0000 &       
 \textbf{5.70} &      
 15.49 &    
 0.0653 &   
 0.0158 &           
 0.06 &          
 17.48 &        
 0.0003 &       
 0.0009 \\

\midrule
\textbf{(Ours)} & \multicolumn{4}{c|}{} & \multicolumn{4}{c|}{} & \multicolumn{4}{c}{} \\

 \textbf{CTXSEG 0.05} &             
 68.00 &              
 0.25 &           
 \textbf{1.0000} &          
 \textbf{0.0770} &      
 24.06 &       
 1.52 &    
 0.9345 &   
 \textbf{0.1196} &          
 24.64 &           
 0.58 &        
 0.9872 &       
 0.1747 \\
 
 \textbf{CTXSEG 0.01} &             
 68.00 &              
 0.26 &           
 \textbf{1.0000} &          
 0.0753 &       
 6.34 &       
 6.49 &    
 0.4837 &   
 0.0875 &          
 11.84 &           
 1.02 &        
 0.9273 &       
 0.2734 \\
 
 \textbf{CTXSEG 0.001} &             
 68.00 &              
 0.27 &           
 \textbf{1.0000} &          
 0.0758 &       
 1.15 &      
 13.48 &    
 0.1262 &   
 0.0273 &           
 \textbf{6.40} &           
 2.05 &        
 0.8170 &       
 \textbf{0.2982} \\
\bottomrule
\addlinespace[3pt]
\multicolumn{13}{p{555pt}}{Comparison between the Varri~\cite{varri_computerized_1988}, non-linear energy operator (NLEO)~\cite{agarwal_adaptive_1999}, Spectral Power Statistics (SPS)~\cite{Jakaite_2011_Feature} and our proposed CTXGEN method, showing count of discovered segment boundaries (target 6), boundary delay (in seconds), sensitivity (range 0-1) and similarity (range 0-1) for each method and signal type. Numbers next to SPS and CTXSEG indicate significance threshold. Each 35 second signal (7 x 5 second segments) contains 6 ground-truth segment boundaries and are generated using harmonics~\cite{Azami_2014_New}, auto-regressive (AR) modeling~\cite{appel_comparative_1984} and our proposed CTXGEN. Results for AR and CTXGEN show an average of 1000 trials to account for stochasticity.}
\end{tabular}
}
\label{tab:result_synthetic}
\end{table*}

\subsection{Experiment 1: Comparing CTXSEG with existing methods using synthetic signals}\label{sec:exp:1}

\revise{R1-4}{In our first experiment, we compare Context Segmentation (CTXSEG) with existing adaptive segmentation methods using synthetic signals.} \revise{R1-6}{We generate synthetic signals using commonly used harmonics and auto-regressive (AR) approaches, as well as our proposed Context Generation (CTXGEN).} In all cases, we generate a 35 second signal with seven 5 second segments, equating to 6 ground-truth segment boundaries. To account for stochasticity in AR and CTXGEN, we generate 1000 signals and report the average. Please refer to~\Cref{appendix:exp1} for specifics.

In \Cref{tab:result_synthetic}, we report the performance of CTXSEG versus the Varri~\cite{varri_computerized_1988}, Non-Linear Energy Operator (NLEO)~\cite{agarwal_adaptive_1999} and Spectral Power Statistics (SPS)~\cite{Jakaite_2011_Feature} methods using their optimal published parameters where appropriate. Since both SPS and our CTXGEN use a statistical distance measure, we report performance across significance thresholds of $\alpha=$ 0.05, 0.01 and 0.001, representing common p-value significance levels for statistical testing in the medical domain~\cite{Chu_1999_introduction,Cesana_2018_What,Ioannidis_2018_Proposal}.

Our proposed CTXSEG method achieves higher boundary similarity scores relative to existing methods, which indicates overall better alignment between discovered and ground-truth segment boundaries after accounting for both over- and under-segmentation. Given appropriate selection of the significance threshold, CTXSEG also demonstrates higher discovery sensitivity and lower incidence of redundant boundaries while keeping boundary delay to a minimum (delay over 5 seconds indicates missed boundary detection, which leads to lower sensitivity). Although the Varri and NLEO methods score highest in terms of detection sensitivity and delay, this comes at a cost of vastly overestimating the number of segment boundaries.

\begin{figure}[ht]
    \centering
    \includegraphics[width=1\linewidth]{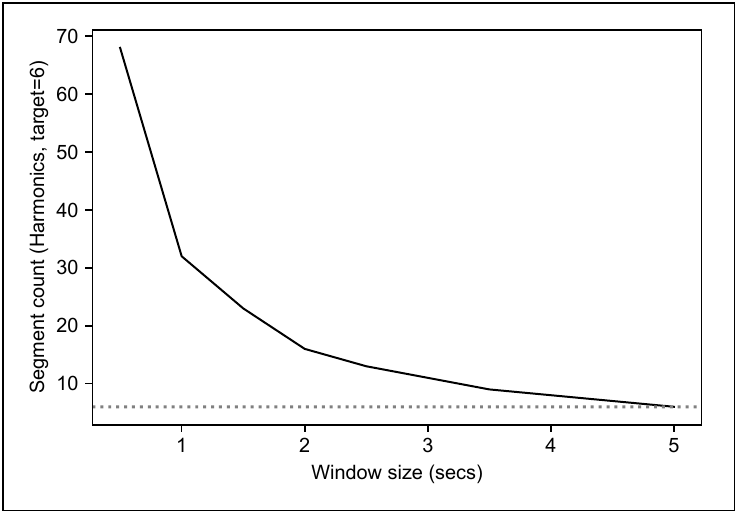}
    \caption{
    Effect of changing the window-size hyperparameter in CTXSEG with relation to slow-moving stationarity changes as governed by the slowest components (0.5 Hz) of the Harmonics signal, where the target is 6 segments (marked with dotted line). Increasing window size reduces over-segmentation and approaches the correct target of 6 segments when window-size is sufficiently large to cover a full period of the slowest component of the Harmonics signal (2 s at 0.5 Hz). This is consistent across different values of significance threshold (alpha).
    }
    \label{fig:slow_moving}
\end{figure}

Of the three signal generation methods, the harmonics signal has relatively slow moving stationarity changes. The high incidence of redundant boundaries for the SPS method confirms that keeping the reference and test windows together (i.e., method of two-contiguous windows) is highly prone to over-segmentation under these circumstances. Although CTXSEG demonstrates more robust and reliable performance by keeping the reference window fixed and sliding only the test window, it is not immune to over segmentation (see~\Cref{fig:slow_moving} and discussion in~\Cref{sec:discussion}).

For AR signals, the source of non-stationarity between segments is driven by models of ictal/non-ictal EEG. For CTXGEN signals, non-stationarity is driven by firing rates of spiking neuron models. In each case, CTXSEG outperforms SPS in all metrics. Interestingly, the ability for SPS to detect segment boundaries is drastically reduced in CTXGEN signals relative to AR, while the our CTXSEG method remains consistent across both signal types. This is reflective of the design choice in the CTXSEG algorithm to compare the full frequency spectrum rather than select frequency bands as in SPS, thereby maintaining general applicability.

\begin{figure*}[!htbp]
    \centering
    \includegraphics[width=1\linewidth]{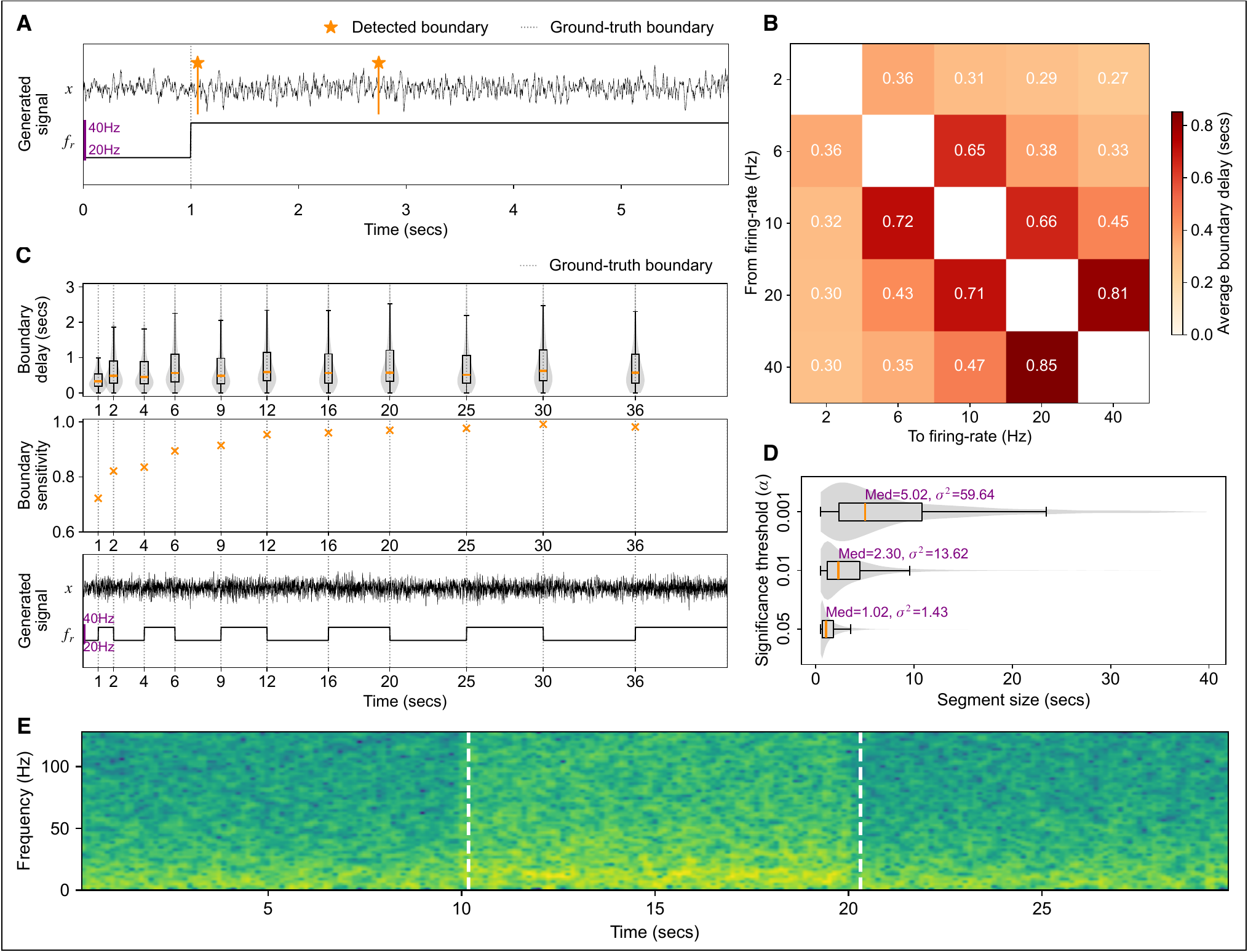}
    \caption{
        Results of Experiment 2. With the exception of signal examples, all results are aggregated from an ensemble of 1000 generated signals. 
        \textbf{A:} Example of a synthetic signal that has been generated using CTXGEN, showing the signal $x$, and step-wise change in firing-rate $f_r$ at $t=1\text{s}$ (20 Hz to 40Hz), which marks the ground-truth segment boundary (dotted line). Two detected segment boundaries, using CTXSEG, are marked with an orange star, with the detection at $t \approx 2.8$ resultant from noise.
        \textbf{B:} Heat map showing the average boundary delay from step-wise changes in firing-rate, where delay is longer with smaller changes in firing-rate. The signals are generated and segmented in the same format as shown in A.
        \textbf{C:} Firing-rates oscillate between 20 Hz and 40 Hz for 1-6 seconds (bottom example), simulating varying context durations. CTXSEG is robust to rapid changes (boundary sensitivity, middle) if context duration exceeds detection time (boundary delay, top).
        \textbf{D:} Distribution of segment-sizes, with median (Med) and variance ($\sigma^2$), for the segments discovered in C, for default significance threshold of $\alpha=0.05$ and more stringent $\alpha \in [0.01, 0.001]$. Segment-sizes are more varied with a more stringent threshold, but also at higher risk of missed boundary detections.
        \textbf{E:} \revise{R1-3}{Spectrogram of a signal generated using CTXGEN with two ground-truth boundaries that switches from firing rate of 2 to 20 Hz at 10 s and 20 back to 2 Hz at 20 s. The boundaries discovered using CTXSEG is marked in white dashed lines. A slight delay is visible when the context states change which corresponds with the expected boundary delay in (B).}
    }
    \label{fig:exp_2}
\end{figure*}

\revise{R1-6}{Our results indicate that CTXGEN signals are particularly challenging for existing segmentation methods, leading to significant over-segmentation by Varri and NLEO, and under-segmentation by SPS. This supports the use of CTXGEN as a versatile test bench for evaluating segmentation algorithms, offering a generally applicable and user-friendly alternative to existing approaches without requiring training data. \track{Here, we refer to ability of the compared algorithms to segment signals in a general capacity and do not draw any conclusions relating to their performance in any specific domain.} We acknowledge that AR-based approaches may be more representative of domain-specific signals.}

\subsection{Experiment 2: Characterizing segmentation ability of CTXSEG using synthetic signals}\label{sec:exp:2}

Having established the segmentation performance of our proposed Context Segmentation (CTXSEG) method in~\Cref{sec:exp:1}, we now seek to understand how the algorithm responds to changes in signal context by generating signals using our proposed Context Generation (CTXGEN) method and measuring boundary delay and sensitivity. Specifically, we investigate the following: 

\begin{enumerate}
    \item First, we assess how changes in signal context, introduced by modifying the firing rates in CTXGEN, affect the time it takes for CTXSEG to detect changes in the signal. We find that smaller changes in the signal result in longer delays in boundary detection.\label{sec:exp:2:item1}
    \item Then, we assess the ability of CTXSEG to respond to rapid changes in the underlying signal. We show that the boundary delay from (item \ref{sec:exp:2:item1}) dictates the minimum duration of changes that CTXSEG can detect, affecting the boundary sensitivity. \label{sec:exp:2:item2}
    \item Finally, we evaluate the impact of the choice of significance threshold on segmentation performance. We find that a more stringent threshold results in longer and more varied segments. \label{sec:exp:2:item3} 
\end{enumerate}

We refer to results reported in~\Cref{fig:exp_2}, with experiment specifics detailed in~\ref{appendix:exp2}. Again, we generate an ensemble of 1000 signals and report the average to account for stochasticity.

\textbf{Characterizing boundary delay:} For item (\ref{sec:exp:2:item1}) above, we generate signals containing pairwise changes across firing rates that are representative of major frequency bands in brain waves. Specifically, we use 2 Hz (to represent delta waves), 6 Hz (theta waves), 10 Hz (alpha waves), 20 Hz (beta waves) and 40 Hz (gamma waves)~\cite{sanei_eeg_2007}. Each signal is 6 seconds long with a single ground truth segment boundary occurring at $t=1$ sec where the firing rate transitions via a step function. 

An example of a synthetic signal that transitions between 20~Hz and 40~Hz is shown in \Cref{fig:exp_2}A, which shows visually complex, random and aperiodic waveforms that are characteristic of real signals. We highlight two discovered boundaries: the first at $t \approx 1.1$ corresponds to a boundary delay of about 0.1 seconds from the ground-truth boundary at $t = 1$, and the second boundary at $t \approx 2.8$ most likely resulted from stochastic noise. 

In \Cref{fig:exp_2}B, we show how boundary delay is affected by the amount of change of firing rate. We find that it is easier to discover segment boundaries (with shorter delay) when changes in signal property are more pronounced. Prior studies of existing methods have found that boundary discovery for downward changes is much more difficult than upward changes \cite{appel_comparative_1984}. While a small difference can still be observed in Figures \ref{fig:exp_2}B and \ref{fig:exp_2}C (top), delay in both directions remains low when considering that \revise{R3-6}{acceptable delay for seizure detection tasks is 10 seconds}~\cite{van_de_vel_automated_2016}.
 
\textbf{\revise{R3-6}{Characterizing boundary sensitivity:}} For item (\ref{sec:exp:2:item2}) above, we show the relationship between boundary delay and boundary sensitivity using a signal that oscillates between two firing rate frequencies, where the period of the oscillation increases from 1 to 6 seconds in 1 second increments. While we investigate the range of firing rates specified in methods for item (\ref{sec:exp:2:item1}) above, we report results for 20~Hz and 40~Hz, as this transition was found to have the longest boundary delay and the most risk of missed detections. A ground truth segment boundary is marked at the point where the firing rate changes. An example of this signal is shown in \Cref{fig:exp_2}C (bottom).

We report the distribution of boundary delay for each ground truth segment boundary across the ensemble in \Cref{fig:exp_2}C (top). We find that a segment boundary can be discovered consistently within 2.5 seconds from the ground-truth boundary irrespective of the actual duration of a stationary period in the signal, with majority occurring within 1 second. This demonstrates that CTXSEG can identify segment boundaries in a timely manner.

For boundary sensitivity, we report the proportion of ground truth segment boundaries that received a boundary discovery before the next ground truth segment boundary and is shown in \Cref{fig:exp_2}C (middle). Our results show that sensitivity exceeds the minimum expected level of 90\% when states persist for longer than the time required to detect a boundary. We base the 90\% benchmark on the expected level of detection sensitivity for seizure detection in the real world~\cite{bruno_seizure_2020}.

\textbf{Impact of significance threshold:} \revise{R2-7}{For item (\ref{sec:exp:2:item3}) above, we use the same signal as in item (\ref{sec:exp:2:item2}) and show the distribution of segment size when using the default threshold $\alpha=0.05$ as well as a more stringent thresholds of $\alpha \in [0.01, 0.001]$ to understand the effect of changing the significance threshold on the segmentation results.}

When using significance threshold $\alpha=0.05$, we find that the median segment size is about 1 second with a variance of 1.43 seconds when the windows size parameter is set to half a second. If CTXSEG were to fail and default to fixed size segmentation, we would expect to see segments with uniform length and zero variance. When we apply the more stringent significance threshold $\alpha \in [0.01, 0.001]$, the segment sizes become longer and more varied. While this may be desirable in certain situations, such as noise handling in long-term EEG, this must be carefully tuned so that segment sizes are not excessively large that they cause state changes to be missed. In our example, a significant portion of segments are longer than the maximum 6 seconds when $\alpha \in [0.01, 0.001]$ but not when $\alpha=0.05$. 

\subsection{Experiment 3: Comparing fixed-length and CTXSEG segmentation methods in seizure detection}\label{sec:exp:3}

In this experiment, we compare fixed-length segmentation to our adaptive Context Segmentation (CTXSEG) in a real-world machine learning (ML) seizure detection task. \track{We find that not only can we use adaptive segmentation with CTXSEG together with existing ML algorithms}, it can improve performance without modifying the underlying machine learning algorithm. 

\track{Specifically, we compare and contrast seizure detection performance by using CTXSEG as a drop-in replacement to fixed-length segmentation in the preprocessing step of a typical EEG machine learning pipeline for seizure detection. We use EEGNet-8,2~\cite{Lawhern_2018_EEGNet} to represent a typical EEG machine learning method. This model has been studied in a variety of EEG machine learning tasks, including seizure detection~\cite{Thuwajit_2022_EEGWaveNet,Zhu_2023_Automated} and brain-computer interface (BCI)~\cite{Lawhern_2018_EEGNet}, as well as used in combination with more advanced models such as Transformers~\cite{Zhu_2023_Automated}.}

\track{For evaluation, we use the standardized SzCORE framework on the widely studied Physionet Siena Scalp EEG dataset~\cite{Detti_2020_SSE} in both the subject specific and subject independent scenarios using the official tools provided by the framework authors~\cite{esl-epfl__szcoreevaluation}. No post-processing (i.e., to reduce false positives) is applied as SzCORE clearly distinguishes samples versus event-based detection. Please refer to~\Cref{appendix:exp3} for detailed experimental specification.}



\begin{table*}[!htbp]
\caption{Comparison between fixed-length and CTXSEG segmentation for seizure detection (subject specific)}
\centering
\scalebox{0.90}{
\begin{tabular}{lcc|rrrr|rrrr}
\toprule
 \multirow{2}{*}{Strategy} & \multirow{2}{*}{Overlap}  & \multirow{2}{*}{Alpha}   & \multicolumn{4}{c|}{Event} &  \multicolumn{4}{c}{Sample} \\
 & & & F1$\uparrow$ & FPR$\downarrow$ & Pres$\uparrow$ & Sens$\uparrow$ & F1$\uparrow$ & FPR$\downarrow$ & Pres$\uparrow$ & Sens$\uparrow$ \\
\midrule
\multirow{3}{*}{Fixed-Fixed} & 25\%               & -               &                    0.1653 &                   187.6500 &                      0.1108 &                      0.9583 &                     0.2738 &                  8,166.6127 &                       0.2534 &                       0.5123 \\
 
 & 50\%               & -               &                    0.1846 &                   164.1909 &                      0.1245 &                      0.9375 &                     0.3043 &                  6,560.6608 &                       0.2748 &                       0.5585 \\
 
 & 75\%               & -               &                    0.1811 &                   171.5304 &                      0.1218 &                      \textbf{0.9896} &                     0.2860 &                  4,298.9210 &                       0.2575 &                       0.6086 \\

\midrule
\textbf{(Ours)} & & & & & & & & & & \\

 \multirow{3}{*}{\textbf{\VarFirst-\VarFirst}}          & -                 & 0.001           &                    0.2010 &                   163.1621 &                      0.1344 &                      0.9688 &                     0.3265 &                  7,821.3060 &                       0.3271 &                       0.5174 \\
 & -                 & 0.01            &                    0.2137 &                   136.4352 &                      0.1438 &                      0.8750 &                     0.3295 &                  7,991.4772 &                       0.3101 &                       0.5197 \\
 & -                 & 0.05            &                    0.2320 &                   144.1981 &                      \textbf{0.1644} &                      0.9271 &                     0.3060 &                  9,455.6636 &                       0.2858 &                       0.5521 \\

 \midrule
 \multirow{3}{*}{\textbf{\VarRand-\VarFirst}} & -                 & 0.001           &                    0.1962 &                   150.0482 &                      0.1306 &                      0.8958 &                     0.3049 &                  8,311.7616 &                       0.2562 &                       0.6065 \\
 & -                 & 0.01            &                    0.1762 &                   170.3704 &                      0.1052 &                      0.9271 &                     0.2845 &                 11,167.9007 &                       0.2352 &                       0.6308 \\
 & -                 & 0.05            &                    0.2370 &                   130.7331 &                      0.1588 &                      0.9062 &                     0.3487 &                  8,615.9362 &                       0.3219 &                       0.5896 \\

\midrule
\multirow{3}{*}{\textbf{\VarFirst-Fixed}}       & 50\%               & 0.001           &                    0.1816 &                   168.8928 &                      0.1243 &                      0.8750 &                     0.2769 &                  8,674.5088 &                       0.2519 &                       0.5235 \\
        & 50\%               & 0.01            &                    0.1821 &                   156.5652 &                      0.1206 &                      0.8958 &                     0.3045 &                 11,824.1133 &                       0.3059 &                       0.5282 \\
        & 50\%               & 0.05            &                    0.1890 &                   169.8273 &                      0.1109 &                      \textbf{0.9896} &                     0.2823 &                 10,181.3248 &                       0.3135 &                       0.4980 \\
 
 \midrule
 \multirow{3}{*}{\textbf{\VarRand-Fixed}}        & 50\%               & 0.001           &                    0.1802 &                   156.0026 &                      0.1091 &                      0.8958 &                     0.2852 &                  9,836.6858 &                       0.2990 &                       0.5480 \\
         & 50\%               & 0.01            &                    0.1550 &                   165.5133 &                      0.0911 &                      0.9062 &                     0.3056 &                  7,396.6539 &                       0.2515 &                       \textbf{0.6320} \\
         & 50\%               & 0.05            &                    0.1352 &                   189.9320 &                      0.0776 &                      0.8958 &                     0.2522 &                 15,323.0363 &                       0.2169 &                       0.6057 \\

 \midrule
 \multirow{3}{*}{\textbf{Fixed-\VarFirst}} & 50\%               & 0.001           &                    \textbf{0.2485} &                    \textbf{98.3559} &                      0.1638 &                      0.8646 &                     \textbf{0.3528} &                  \textbf{2,885.8907} &                       \textbf{0.3555} &                       0.4521 \\
 & 50\%               & 0.01            &                    0.1931 &                   105.6779 &                      0.1151 &                      0.8542 &                     0.3274 &                  6,232.3050 &                       0.3100 &                       0.4911 \\
 & 50\%               & 0.05            &                    0.2271 &                   113.3906 &                      0.1510 &                      0.8958 &                     0.3346 &                  3,630.4410 &                       0.2984 &                       0.5461 \\
\bottomrule
\addlinespace[3pt]
\multicolumn{11}{p{460pt}}{Test set evaluation showing F1, false positive rate (FPR), precision (Pres) and sensitivity (Sens), between fixed segmentation and variable CTXSEG strategies for seizure detection on the Siena Scalp EEG dataset using SzCORE evaluation. Tests are conducted using EEGNet-8,2 with identical parameters and only changing the segmentation method in preprocessing. Overlap is only applicable with Fixed segmentation and Alpha is only applicable with CTXSEG segmentation (i.e. \VarFirst, \VarRand).}
\end{tabular}
}
\label{tab:ctxseg_sse_ss_agg}
\end{table*}
\begin{table*}[!htbp]
\caption{Comparison between fixed-length and CTXSEG segmentation for seizure detection (subject independent)}
\centering
\scalebox{0.9}{
\begin{tabular}{lcc|rrrr|rrrr}
\toprule
 \multirow{2}{*}{Strategy} & \multirow{2}{*}{Overlap}  & \multirow{2}{*}{Alpha}   & \multicolumn{4}{c|}{Event} &  \multicolumn{4}{c}{Sample} \\
 & & & F1$\uparrow$ & FPR$\downarrow$ & Pres$\uparrow$ & Sens$\uparrow$ & F1$\uparrow$ & FPR$\downarrow$ & Pres$\uparrow$ & Sens$\uparrow$ \\
\midrule
\multirow{3}{*}{Fixed-Fixed}              & 25\%               & -               &                    0.3303 &                    60.0103 &                      0.2909 &                      0.7310 &                     0.2519 &                    927.4341 &                       0.4255 &                       0.3056 \\
               & 50\%               & -               &                    0.3139 &                    56.1711 &                      0.2981 &                      0.6000 &                     0.2213 &                  1,432.5078 &                       0.4419 &                       0.2683 \\
               & 75\%               & -               &                    0.3053 &                    57.8782 &                      0.2678 &                      0.6905 &                     0.2422 &                    783.7437 &                       0.4197 &                       0.2531 \\

\midrule
\textbf{(Ours)} & & & & & & & & & & \\

\multirow{3}{*}{\textbf{\VarFirst-\VarFirst}}          & -                 & 0.001           &                    0.3301 &                    57.7862 &                      0.2685 &                      0.7833 &                     0.2589 &                    783.5286 &                       0.3908 &                       0.2797 \\
           & -                 & 0.01            &                    0.3144 &                    65.6839 &                      0.2919 &                      0.7167 &                     0.2071 &                    982.8682 &                       0.3944 &                       0.2612 \\
           & -                 & 0.05            &                    0.2890 &                    61.3200 &                      0.2360 &                      0.6857 &                     0.1868 &                  1,265.5910 &                       0.3192 &                       0.2527 \\

\midrule
\multirow{3}{*}{\textbf{\VarRand-\VarFirst}}          & -                 & 0.001           &                    0.3379 &                    59.9380 &                      \textbf{0.3239} &                      0.6976 &                     0.2213 &                  1,009.1394 &                       0.3802 &                       0.2743 \\
           & -                 & 0.01            &                    0.3383 &                    \textbf{51.9872} &                      0.2879 &                      0.6976 &                     0.2180 &                    705.9719 &                       0.3804 &                       0.2104 \\
           & -                 & 0.05            &                    0.3272 &                    65.4239 &                      0.2731 &                      0.8190 &                     0.2555 &                    865.9221 &                       0.3960 &                       0.2710 \\

\midrule
\multirow{3}{*}{\textbf{\VarFirst-Fixed}}       & 50\%               & 0.001           &                    \textbf{0.3727} &                    59.6015 &                      0.3183 &                      0.7738 &                     0.2386 &                    945.3488 &                       0.4644 &                       0.2696 \\
        & 50\%               & 0.01            &                    0.3591 &                    66.0863 &                      0.2926 &                      \textbf{0.8762} &                     0.2603 &                  1,063.2493 &                       \textbf{0.4739} &                       0.2871 \\
        & 50\%               & 0.05            &                    0.3351 &                    66.7451 &                      0.2695 &                      0.8619 &                     \textbf{0.2767} &                    \textbf{620.1964} &                       0.4387 &                       0.3088 \\

 \midrule
 \multirow{3}{*}{\textbf{\VarRand-Fixed}}        & 50\%               & 0.001           &                    0.3329 &                    68.5456 &                      0.2810 &                      0.8143 &                     0.2654 &                    922.6022 &                       0.4069 &                       0.2878 \\
         & 50\%               & 0.01            &                    0.2999 &                    61.0121 &                      0.2624 &                      0.6833 &                     0.1896 &                    727.9335 &                       0.3952 &                       0.2078 \\
         & 50\%               & 0.05            &                    0.2356 &                    79.8149 &                      0.2116 &                      0.7310 &                     0.2365 &                  1,135.8871 &                       0.3732 &                       0.2992 \\

\midrule
\multirow{3}{*}{\textbf{Fixed-\VarFirst}}          & 50\%               & 0.001           &                    0.3359 &                    63.3634 &                      0.2974 &                      0.7381 &                     0.2594 &                    987.8630 &                       0.4278 &                       0.2877 \\
          & 50\%               & 0.01            &                    0.3430 &                    58.5452 &                      0.3019 &                      0.7476 &                     0.2692 &                    913.4611 &                       0.4267 &                       \textbf{0.3143} \\
          & 50\%               & 0.05            &                    0.3203 &                    53.7749 &                      0.2732 &                      0.7262 &                     0.2369 &                    722.7299 &                       0.4183 &                       0.2421 \\
\bottomrule
\addlinespace[3pt]
\multicolumn{11}{p{460pt}}{Test set evaluation showing F1, false positive rate (FPR), precision (Pres) and sensitivity (Sens), between fixed segmentation and variable CTXSEG strategies for seizure detection on the Siena Scalp EEG dataset using SzCORE evaluation. Tests are conducted using EEGNet-8,2 with identical parameters and only changing the segmentation method in preprocessing. Overlap is only applicable with Fixed segmentation and Alpha is only applicable with CTXSEG segmentation (i.e. \VarFirst, \VarRand)}
\end{tabular}
}
\label{tab:ctxseg_sse_si}
\end{table*}

\textbf{Subject specific performance:} We report subject-specific performance by first averaging time-series cross-validation (TSCV) performance within subjects, then between subjects (\Cref{tab:ctxseg_sse_ss_agg}). All CTXSEG strategies show improvements over fixed segmentation in most metrics. Most notably are the substantial reductions in event false positive rate (FPR) leading to improvements in event-based F1.

Of the three proposed CTXSEG strategies, Fixed-\VarFirst\ shows most consistent improvements over pure fixed segmentation strategies. This is also the most versatile of the strategies given that CTXSEG is applied only at test time. When CTXSEG is applied for training, selecting the first window of each segment (\VarFirst-\VarFirst) is marginally more robust than selecting a random window (\VarRand-\VarFirst) (more on this in~\Cref{sec:discussion}).

\begin{figure*}
    \centering
    \includegraphics[width=1\linewidth]{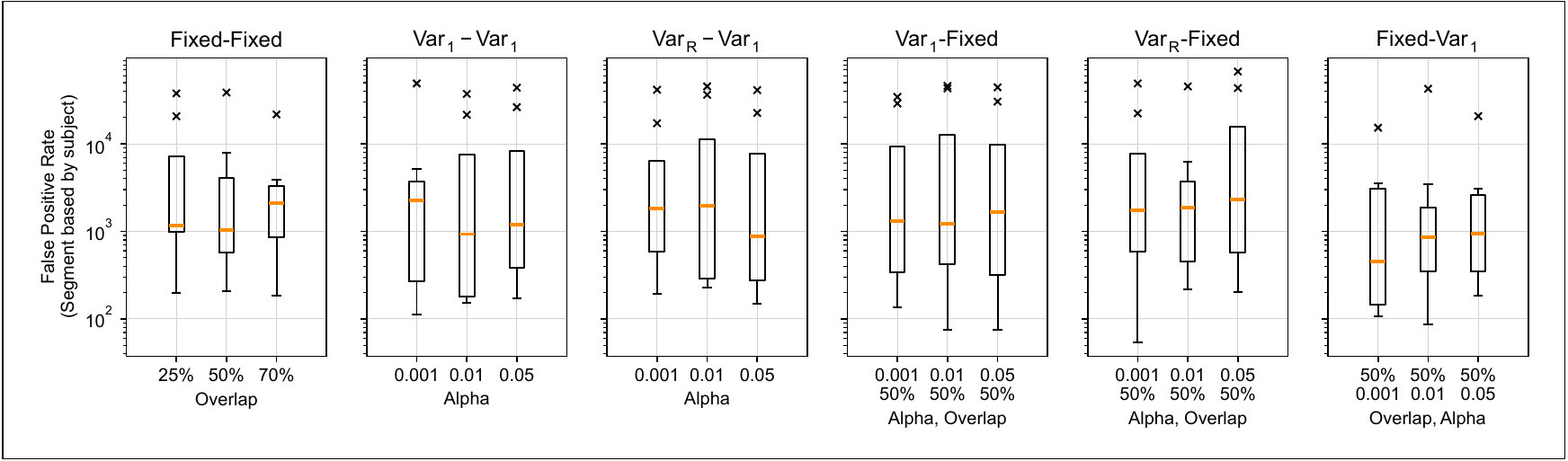}
    \caption{
    Comparison of segmentation strategies based on the distribution of sample based false positive rate (FPR) in subject specific evaluation with SzCORE on the Siena Scalp EEG dataset using the EEGNet-8,2 model. FPR distribution improves (decreases) when using CTXSEG at test time (Fixed-\VarFirst, training with fixed-sized segments at 50\% overlap and applying CTXSEG during inference). However, there is increased variance in FPR when training using variable-length segments. Distribution outliers are indicated by a cross.
    }
    \label{fig:exp_3_fpr}
\end{figure*}

CTXSEG is shown to generally reduce event-based FPR; however, sample FPR is surprisingly high for strategies using CTXSEG for training. We think that since SzCORE scores sample-based metrics at 1 Hz resolution, false predictions that are concentrated chronologically can result in high sample FPR but low event FPR. In~\Cref{fig:exp_3_fpr}, we also see that, while median sample FPR is low, there is more variance between subjects when using CTXSEG for training. We attribute this to the lower training segment count when using CTXSEG (see~\Cref{tab:ctxseg_seg_stats}). The same phenomenon is not seen when CTXSEG is used only at test time (Fixed-\VarFirst).

\textbf{Subject independent performance:} We obtain subject-independent performance by averaging leave-one-out cross-validation (LOOCV) across subjects (\Cref{tab:ctxseg_sse_si}). Here, performance difference between segmentation strategies is less pronounced. Nevertheless, segmentation with CTXSEG offers modest performance improvements in most metrics, with strategies \VarRand-\VarFirst, \VarFirst-Fixed most favorable.

Overall, we observe better metric performance with subject-independent evaluation than with subject-specific evaluation. This is surprising given that cross-subject seizure detection is usually expected to be a more difficult task. We attribute this to the larger volume of data available for training, with each LOOCV fold in subject-independent evaluation encompassing the entire dataset minus one subject (see~\Cref{tab:ctxseg_seg_stats}). In contrast, some TSCV folds for subject-specific evaluation may only contain as few as one EEG record and one labeled seizure event. \track{Note that while the number of segments for training can vary due to the segmentation method, the overall data coverage is the same across all experiments.}

\textbf{Performance benefits:} We attribute the performance benefits from CTXSEG, especially when applied at test time (as in Fixed-\VarFirst), to lower segment counts resulting from larger and more varied segment sizes. This allows a classifier to focus on select portions of the EEG where the signal properties change (more on this in~\Cref{sec:discussion}). \revise{R3-6}{The most significant concern with this approach is whether the discovery of segments is sufficiently robust so that events of interest that occur for a short duration, such as seizures, are not missed and absorbed into background segments. We investigate the impact of this concern by considering the boundary delay.} 

\begin{figure}[t]
    \centering
    \includegraphics[width=1\linewidth]{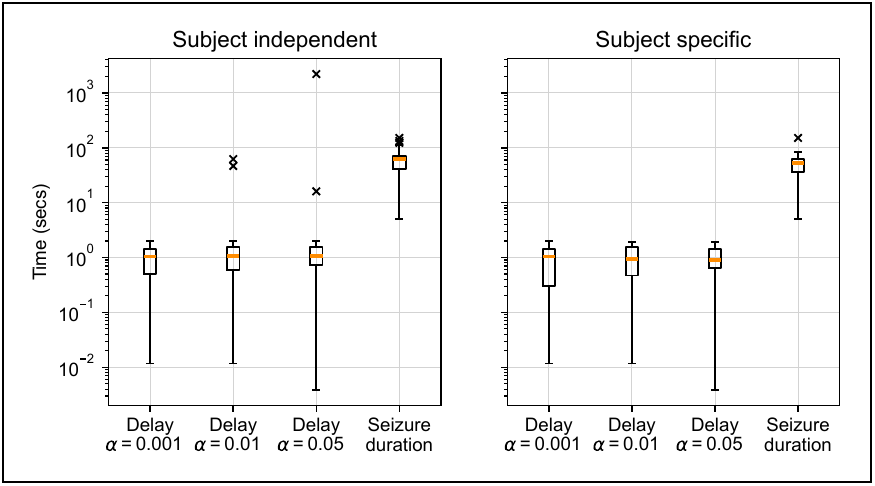}
    \caption{
    Comparison of distributions of boundary delay relative to seizure duration when using CTXSEG across different values of Alpha. There is risk of missed detections (leading to lower detection sensitivity) if boundary delay exceeds seizure duration. Distribution outliers are indicated by a cross.
    }
    \label{fig:exp_3_delay}
\end{figure}

\textbf{\revise{R3-6}{Boundary delay and sensitivity:}} Boundary delay represents the earliest opportunity available for a classifier to classify a segment that is labeled for the seizure class, and is not to be confused with the time taken by the classifier to detect a seizure.

We report boundary delay in~\Cref{fig:exp_3_delay}. Median boundary delay is approximately 1 second, which is well within the duration of seizures lasting 8-100 seconds, providing ample opportunity for a classifier to detect the seizure. Boundary delay is also mostly within the acceptable 10 second range for seizure detection in clinical settings~\cite{van_de_vel_automated_2016}. Boundary delay does exceed seizure duration in isolated cases, which may lead to the detection of these events being missed. However, this is not observed when $\alpha=0.001$, which suggests that such risks can be mitigated by tuning CTXSEG parameters specific to the task and data set. 

\begin{table*}[!htbp]
\caption{Comparative segment statistics between fixed-length and CTXSEG segmentation}
\centering
\scalebox{0.9}{
\begin{tabular}{cccc|rr|rrr}
\toprule
\multirow{2}{*}{Scenario} & \multirow{2}{*}{Segmentation}  & \multirow{2}{*}{Overlap}   & \multirow{2}{*}{Alpha} & \multicolumn{2}{c|}{Segment size} &  \multicolumn{3}{c}{Segment count} \\
 & & & & Mean & Std dev & Train & Dev & Test \\
\midrule
\multirow{6}{*}{\makecell{Subject\\specific}}    & \multirow{3}{*}{Fixed}                  & 25\%               & -               &                   512.0000 &                        0.0000 &                   9,590.1875 &                 6,481.8125 &                  7,049.1250 \\
     &                   & 50\%               & -               &                   512.0000 &                        0.0000 &                  14,384.1875 &                 9,722.1250 &                 10,573.1250 \\
     &                   & 75\%               & -               &                   512.0000 &                        0.0000 &                  28,768.3750 &                19,444.2500 &                 21,146.2500 \\
     \cmidrule{2-9}
     & \multirow{3}{*}{CTXSEG} & -                 & 0.001           &                   520.6703 &                      173.1654 &                   7,068.7500 &                 4,766.1250 &                  5,159.8750 \\
     &   & -                 & 0.01            &                   516.1919 &                       83.5787 &                   7,117.8125 &                 4,817.0625 &                  5,233.3125 \\
     &    & -                 & 0.05            &                   516.0944 &                      144.2121 &                   7,074.6875 &                 4,829.5625 &                  5,248.0000 \\

 \midrule
 \multirow{6}{*}{\makecell{Subject\\independent}} & \multirow{3}{*}{Fixed}                  & 25\%               & -               &                   512.0000 &                        0.0000 &                 290,064.8571 &                24,172.0714 &                 24,172.0714 \\
  &                   & 50\%               & -               &                   512.0000 &                        0.0000 &                 435,078.8571 &                36,256.5714 &                 36,256.5714 \\
  &                   & 75\%               & -               &                   512.0000 &                        0.0000 &                 870,157.7143 &                72,513.1429 &                 72,513.1429 \\

  \cmidrule{2-9}
  & \multirow{3}{*}{CTXSEG} & -                 & 0.001           &                   556.1894 &                    1,277.6331 &                 200,221.7143 &                16,685.1429 &                 16,685.1429 \\
  &                 & -                 & 0.01            &                   540.8601 &                    1,057.9044 &                 205,948.2857 &                17,162.3571 &                 17,162.3571 \\
  &                 & -                 & 0.05            &                   537.0651 &                    1,053.4611 &                 207,362.5714 &                17,280.2143 &                 17,280.2143 \\

\bottomrule
\addlinespace[3pt]
\multicolumn{9}{p{450pt}}{Aggregated segment statistics across train/dev/test sets for Experiment 3, under the SzCORE evaluation framework, created using either fixed for adaptive CTXSEG segmentation and grouped by either subject specific or subject independent scenarios. Subject specific scenarios typically have much lower segment counts for training. In contrast subject independent utilize the entire dataset minus one patient. Adaptive segmentation with CTXSEG typically creates longer and more varied segment sizes.}
\end{tabular}
}
\label{tab:ctxseg_seg_stats}
\end{table*}

\section{Discussion}\label{sec:discussion}

Analysis of EEG with modern machine learning techniques relies on dividing the long-duration continuous signals into manageable fixed-length segments. Currently, the status quo is to arbitrarily select fixed-length segments at fixed time intervals without consideration of the underlying signal activity. We have empirically shown that segmenting EEG adaptively, that is considerate of signal activity, is both compatible with modern machine learning techniques and may be more robust than naive fixed-length methods under certain circumstances. \track{Since ground-truth segment boundaries are generally lacking in most EEG datasets, we have focused on the net impact to EEG machine learning tasks (i.e. seizure detection) rather than the accuracy of the discovered segment boundaries.}

\textbf{\revise{R1-1}{Strengths:}} Applying our proposed Context Segmentation (CTXSEG) to an EEG seizure detection task has shown that adaptive segmentation can be most useful when applied at test time (see~\Cref{sec:exp:3}). We identify that a key advantage of segmenting adaptively lies with the observation that segments are temporally longer, yielding fewer segments overall (\Cref{tab:ctxseg_seg_stats}). This allows a classifier to focus on the most notable temporal positions in the EEG, thereby reducing opportunity for noisy predictions. This is clearly supported by the lower incidence of false positive rate (FPR), higher precision and higher F1 scores in our experimental results. Since lower segment count may also lead to more efficient inference, it may be interesting to investigate possible efficiency benefits for EEG machine learning in online applications in future work.

\textbf{\revise{E-4,R1-1}{Limitations and future improvements:}} We observe that using CTXSEG for model training could be less robust than fixed-length methods in terms of detection sensitivity under certain circumstances. This is observed with subject-specific (\Cref{tab:ctxseg_sse_ss_agg}) but not subject-independent (\Cref{tab:ctxseg_sse_si}) evaluation. Since CTXSEG yields fewer segments for training, there is potential for under-sampling of the training data. This could be compounded by the lower overall proportion of segments available for training in the subject-specific scenario under the SzCORE framework (\Cref{tab:ctxseg_seg_stats}). Our results also show that selecting the first window to represent a variable-length segment is generally correlated with higher sensitivity, although it may not always lead to stronger overall F1 performance. There are further research opportunities here to find the best fixed-length representation of these variable-length segments. 

\track{We acknowledge that while our heuristic approach in CTXSEG, relating to statistical similarity of frequency characteristics, is intended for general application, there may be alternative approaches that may be more suitable on an application-specific basis. In addition, our design of CTXSEG to operate linearly over the signal in a single pass favors computational efficiency over segmentation accuracy. Since segment boundaries are discovered on a greedy basis with no opportunity for refinement, the algorithm is sensitive to local maxima with respect to context changes. There is also opportunity for future work to explore methods that refine boundaries while keeping computational costs to a minimum.}

\track{In addition, real EEG is typically noisy and may contain slow stationarity changes that evolve over longer time durations, both factors that could confound our ability to make accurate discovery of segment boundaries. Noise in EEG can originate from a variety of sources including electrical interference, muscle activity and/or other biological processes. We note that our current implementation of CTXSEG does not have built-in mechanisms to suppress noise and is a possible avenue for improvement in future work. Although this is the case, the window size hyperparameter in CTXSEG directly affects the resolution of the power spectra used for statistical comparison, where a larger window size could be expected to promote resilience towards noisy signals. The impact of noise may be estimated by inspecting the distribution of segment size and measuring variance as shown in \Cref{fig:exp_2}D.}

\track{For long-duration stationarity changes, we show that while CTXSEG is more robust than traditional adaptive segmentation methods (\Cref{tab:result_synthetic}), it is still susceptible to over-segmentation when the period of slow-moving components exceeds that of window size (see \Cref{tab:result_synthetic}). In particular, we note that window size may need to be set at a duration that is much longer than the period of the slowest moving frequency component in order to be robust to that component (\Cref{fig:slow_moving}).}

\textbf{\revise{R2-7}{Tuning hyperparameters:}} There are three main hyperparameters for CTXSEG: window size, stride and significance threshold ($\alpha$). We have shown that increasing window size can be favorable to capturing slow moving stationarity changes (\Cref{fig:slow_moving}). Larger window sizes may also be favorable if the underlying signal is noisy. Since window size dictates the size of the smallest discoverable segment, this needs to be carefully balanced with events of interest (such as seizures) that may have relatively short duration, where excessively large window sizes may lead to segment boundaries being missed and lower detection sensitivity.

We have also shown that lower values of alpha promote the discovery of longer and more varied segments (\Cref{fig:exp_2}D and \Cref{tab:ctxseg_seg_stats}). Although there are observable trends relating to alpha when segmenting synthetic signals (\Cref{tab:result_synthetic}), seizure detection performance using real EEG remains mixed (Tables \ref{tab:ctxseg_sse_ss_agg} and \ref{tab:ctxseg_sse_si}), with the optimal value of alpha likely application- and dataset-specific. For all of our experiments, we use stride of one sample to maximize boundary detection resolution.

\section{Conclusion}\label{sec:conclusion}

\track{EEG preprocessing for machine learning often involves fixed-length segmentation, however, this may have limited biologically relevance. Adaptive segmentation creates variable-length segments that more aligned with signal activity, however, they may not be compatible with modern machine learning approaches that require fixed-size input.}

\track{In this work, we revisit adaptive segmentation as a viable preprocessing method for EEG analysis using machine learning. We propose novel methods for adaptive segmentation and for using variable-length segments in machine learning algorithms that typically require fixed-length input.}

\track{Our proposed method, CTXSEG, can be used as a drop-in replacement for fixed-length segmentation in modern machine learning methods and has potential to improve performance with standardized benchmarks, even when applied only at test time without needing additional training. This has the added benefit of requiring fewer segments, potentially improving efficiency for online applications.}

\track{We recommend that adaptive segmentation be considered as part of the standard preprocessing repertoire in EEG machine learning applications. Future work could explore more robust methods for segmenting the EEG, refining segment boundaries, suppressing the discovery of noisy segments, representing variable-length EEG segments for machine learning, and assessing potential efficiency improvements for online EEG applications.}

\section*{Conflict of interest statement}

The authors do not have any conflicts of interest in relation to this work.


\section*{Data access statement}

The publicly accessible Siena Scalp EEG Database \cite{Detti_2020_SSE} can be obtained from the following link \hyperlink{https://doi.org/10.13026/5D4A-J060}{https://doi.org/10.13026/5D4A-J060}, and SzCORE BIDS compatible version can be accessed from following link \hyperlink{http://doi.org/10.5281/ZENODO.10640761}{http://doi.org/10.5281/ZENODO.10640761}. The publicly accessible Bonn University dataset \cite{andrzejak_indications_2001} can be obtained from the following link \hyperlink{http://hdl.handle.net/10230/42894}{http://hdl.handle.net/10230/42894}.



\appendix

\appendices
\label{appendix}

\section{Existing adaptive segmentation methods}\label{sec:background:adaptive_segmentation}


\textbf{Algorithmic overview:} There are two main components common to all existing adaptive segmentation methods:  an algorithm for sliding-windows and a measure of comparison for these windows. 

\textit{Sliding-window} algorithms make use of two windows, typically referred to as the \textit{reference} and \textit{test} windows, that slide along the signals with the reference window acting as the baseline template to which the test window is compared. The window size is usually shorter than the shortest expected segment but long enough to reveal the slowest frequency components \cite{michael_automatic_1979}.
Comparison between reference and test windows is facilitated by a \textit{distance measure}, which quantifies the degree of stationarity between the two windows. Segment boundaries are discovered when the distance measure exceeds a user-defined threshold, which indicates that a significant change in signal properties has occurred and the signal is no longer stationary. The distance measure can be parametric or non-parametric.

\textbf{Parametric distance measures:} Methods using parametric distance measures typically rely on modeling the EEG as auto-regressive (AR) processes. A segment is considered quasi-stationary if the parameters of the AR model remain constant \cite{aufrichtigl_adaptive_1991}. AR models follow the general form $x_t = \sum_{i=1}^{\rho}{\theta_i  x_{t-i} + \epsilon_t}$ with parameters $\theta$, order $\rho$, and white noise $\epsilon$.

\textit{Spectral Error Measure} (SEM) \cite{bodenstein_feature_1977} is a distance measure based on the amount of prediction error made by an auto-regressive linear predictor \cite{makhoul_linear_1975}. The algorithm compares the distance measure between a fixed reference window and a test window that slides forward in time, both of which are equal in size. When a segment boundary is discovered, a new reference window is established at the boundary point and the algorithm is repeated.

\textit{Generalized likelihood ratio} (GLR) \cite{appel_adaptive_1983} is a distance measure that tests the likelihood that parameters of the reference and test windows are the same when modeled as AR-processes. The sliding-window algorithm functions similarly to the SEM method, except that the reference window grows in size as the test window slides forward in time.

\textbf{Non-parametric distance measures:} Methods that use non-parametric distance measures are often considered advantageous as they do not require estimation of parameters from EEG \cite{e_brodsky_nonparametric_1999,oppenheim_discrete-time_1999}. 

The \textit{autocorrelation function} (ACF) method is one of the first non-parametric methods for adaptive segmentation \cite{michael_automatic_1979}. This method uses a distance measure based on changes in amplitude and frequency relative to predefined thresholds that is estimated from the autocorrelation functions of the reference and test windows. The algorithm for sliding-windows is identical to that of the SEM-method.

\textit{Two-contiguous windows} is a family of methods that was first proposed by \cite{silin_automatic_1986}. A common feature of these methods is the algorithm for sliding windows that makes use of equally sized reference and test windows that are contiguous and slide forward in unison. This algorithm is considered advantageous in an online environment \cite{krajca_online_1991} that required simultaneous and parallel segmentation of EEG with multiple channels. Since stationarity can occur at different times in different channels, it was seen as problematic to keep the reference and test windows synchronized in the SEM and GLR based sliding windows algorithms.

The distance measure $G$ used in the two-contiguous windows method, as originally proposed by \cite{silin_automatic_1986,krajca_automatic_1991}, compares the maximal difference in the power spectra between the reference window $X(\omega)$ and the test window $Y(\omega)$, both computed using the Fast Fourier Transform (FFT). Local maxima, rather than a threshold, is used to identify segment boundaries,
\begin{equation}
    G = \max_{\omega} \left\{ \frac{1}{2} \left[ \frac{X(\omega)}{Y(\omega)} + \frac{Y(\omega)}{X(\omega)} \right] - 1 \right\}
\end{equation}

While popular due to the ease of online processing of multi-channel EEG-signals, computing the FFT for every time step in a signal was considered to be computationally intensive in the context of the limited computing resources at the time (circa 1988) \cite{varri_computerized_1988,krajca_automatic_1991}. The \revise{R1-4}{Varri method}~\cite{varri_computerized_1988} replaces the FFT with frequency ($\text{FDIF}$) and amplitude ($\text{ADIF}$) estimates. For a window of size $w$ and signal $\mathbf{x}$:
\begin{align}
    \text{FDIF} &= \sum_{i=1}^{w-1} \vert x_i - x_{i-1} \vert,\\
    \text{ADIF} &= \sum_{i=1}^{w-1} \vert x_i \vert.
\end{align}

The distance $G$ between the first and second windows is then calculated using empirically defined coefficients $k_A = 1$ and $k_F = 7$~\cite{krajca_automatic_1991},
\begin{equation}
    G = k_A \vert \text{ADIF}_1 - \text{ADIF}_2 \vert + k_F \vert \text{FDIF}_1 - \text{FDIF}_2 \vert
\end{equation}

The method was highly prone to the discovery of redundant segment boundaries and was less sensitive to slow changes (trends) with duration longer than the window sizes. One method to remedy this shortcoming is to use an adaptive threshold ($\text{THR}$) to reject insignificant local maxima~\cite{krajca_automatic_1991}. For a block of signal with $\text{BL}$ samples,
\begin{equation}
    \text{THR} = \frac{1}{\text{BL}}(k_A \text{ADIF} + k_F \text{FDIF})
\end{equation}

Alternative measurements of distance, such as \revise{R1-4}{non-linear energy operators (NLEO)}~\cite{agarwal_adaptive_1999}, fractal dimensions \cite{kirlangic_fractal_2001,Azami_2012_New} and evolutionary algorithms~\cite{Azami_2013_hybrid,Azami_2015_intelligent}, have also been explored. In the case of NLEO~\cite{agarwal_adaptive_1999}, the energy $Q$ of the signal at position $n$ can be expressed as
\begin{equation}
    Q(n) = x_{n-1} x_{n-2} - x_{n} x_{n-3}.
\end{equation}

The distance $G_{\text{NLEO}}$ at position $n$ is calculated as the absolute difference between energy in left and right windows, each of size $w$~\cite{agarwal_adaptive_1999},
\begin{equation}
    G_{\text{NLEO}}(n) = \left\vert \sum_{i=n-w+1}^{n} Q(i) - \sum_{i=n+1}^{n + w} Q(i) \right\vert.
\end{equation}

\revise{R1-4}{Spectral Power Statistics (SPS)}~\cite{Jakaite_2011_Feature} uses two-sample hypothesis testing for the distance measure with a user-defined threshold. This compares the spectral power sums for nine frequency bands between the two contiguous windows that is obtained from the discrete Fourier transform. When compared to the chi-squared, t-test and Kolmogorov-Smirnov tests, the Anderson-Darling test was found to be most informative in terms of correlating newborn sleep stage EEG with age.

\section{Experiment 1 Supplementary}\label{appendix:exp1}

\textbf{Harmonics signal:} Harmonics-based synthetic signals require manual definition of ground-truth segments that enact changes in amplitude and/or frequency. Although simple, parameters are often arbitrarily determined and may not accurately represent the complexity and variability of real signals. An example of a common harmonics pattern with seven ground-truth segments from~\cite{Azami_2014_New} is
\begin{equation}\label{eq:harmonics}
    \begin{aligned}
        &0.5 \cos(\pi t) + 1.5\cos(4 \pi t) + 4 \cos(5 \pi t),\\
        &0.7 \cos(\pi t) + 2.1 \cos(4 \pi t) + 5.6 \cos(5 \pi t),\\
        &1.5\cos(2 \pi t) + 4 \cos(8 \pi t),\\
        &1.5\cos(\pi t) + 4 \cos(4 \pi t),\\
        &0.5 \cos(\pi t) + 1.7 \cos(2 \pi t) + 3.7 \cos(5 \pi t),\\
        &2.3 \cos(3 \pi t) + 7.8 \cos(8 \pi t),\\
        &0.8 \cos(\pi t) + \cos(3 \pi t) + 3 \cos(5 \pi t).
    \end{aligned}
\end{equation}  

We generate a synthetic signal, referred to as \textit{Harmonics} using~\Cref{eq:harmonics}. Each component (total 7) within the equation relates to a single segment lasting 5 seconds, with the total signal 35 seconds in duration.

\textbf{Auto-regressive signals:} Auto-regressive (AR) approaches involve fitting multiple AR-models to real EEG, making them more robust at capturing the characteristics of real signals than harmonics. Although signals generated using AR-models are more complex, representativeness of real signals remains in question~\cite{wendling_segmentation_1997}. Additionally, the manual nature of this approach means that only a limited number of configurations can be examined, affecting its overall versatility.

We generate synthetic signals referred to as \textit{Auto-regressive} by using the method described in~\cite{appel_comparative_1984} and use two models that correspond to non-ictal/ictal states of a real EEG dataset. We use the Bonn University EEG dataset~\cite{andrzejak_indications_2001} and train the non-ictal models using set ``A" and the ictal model using set ``E". When generating the signals, we use $7 \times 5$ second segments corresponding to the following sequence $\{ \text{A}, \text{E}, \text{A}, \text{E}, \text{A}, \text{E}, \text{A} \}$. To generate one signal, we randomly select one representative record from each sets A and E to fit the models before proceeding with signal generation. To account for random selection, we generate 1000 signals (referred to as trials) for our experiments.

\textbf{CTXGEN signals:} We generate synthetic signals using our proposed Context Generation (CTXGEN) method. We use $7 \times 5$ second segments with firing rates (in Hz) corresponding to the following sequence $\{ 2, 40, 20, 10, 40, 6, 20 \}$ and $m = 500$ being the number of spiking neuron models used in the generation process. To account for random selection, we generate 1000 signals (referred to as trials) for our experiments.

\textbf{Segmentation parameters:} We use sampling frequency of 256 Hz and window size of 0.5 seconds and stride of 1 sample across all segmentation methods. For the Varri method~\cite{varri_computerized_1988}, we additionally use window size of 8 seconds for the adaptive threshold and 0.1 seconds for calculating local minima~\cite{krajca_automatic_1991}. For the non-linear energy operator (NLEO) method~\cite{agarwal_adaptive_1999}, we also use 0.1 seconds for calculating local minima. For Spectral Power Statistics (SPS)~\cite{Jakaite_2011_Feature}, we use the Anderson-Darling test for statistical testing.

\section{Experiment 2 Supplementary}\label{appendix:exp2}

Each synthetic signal in this experiment is generated using Context Generation (CTXGEN) at a sampling frequency of 256 Hz. Signals are segmented using Context Segmentation (CTXSEG) with window size $w = 128$ samples (0.5 seconds), stride $s=1$ sample and significance threshold $\alpha = 0.05$. We apply a Hamming window is applied to both the reference and test windows and compare the log magnitude of the FFT is compared using the paired t-test. To account for random noise, we generate an ensemble of 1000 signals for each firing rate pair and report the mean.

\section{Experiment 3 Supplementary}\label{appendix:exp3}

\textbf{Seizure classifier:} We base our experiments on the well-known EEGNet-8,2 model~\cite{Lawhern_2018_EEGNet} and use the optimal settings as described in the original publication, with the exception that we retain the input sampling frequency at 256 Hz (instead of 128 Hz) and adjust the kernel size accordingly to 128 (from 64).

\textbf{SzCORE evaluation:} We evaluate with the SzCORE framework in both the subject-specific and subject-independent scenarios using the official evaluation tools provided by the framework authors~\cite{esl-epfl__szcoreevaluation}. 

\textbf{Dataset:} We use the Physionet Siena Scalp EEG dataset~\cite{Detti_2020_SSE}, which is a commonly studied dataset that is supported by the SzCORE framework. We use the BIDS compatible version provided by the framework authors~\cite{Dan_2024_SSE_BIDS}, which is standardized for 19 channels at a sampling frequency of 256 Hz. For subject-independent evaluation, all subjects and EEG records (referred to as runs) are used. Data for subjects ``sub-03", ``sub-07", ``sub-11", ``sub-16" and ``sub-17", do not meet the minimum requirement of having 3 seizure events and so were excluded from subject-specific evaluation. A manifest of subjects and runs used for the train/dev/test sets in each of the cross-validation folds is provided as part of the supplementary material~\cite{Zhou_2025_SSE_BIDS_Manifest}. 

\textbf{Preprocessing:} We apply a bandpass filter between 1.5-40 Hz that is consistent with EEGNet-8,2 methodology, with a slight variation being the use of 1.5 Hz rather than 1 Hz on the lower bound to eliminate heart rate effects visible for some subjects. Filtered data is then scaled by removing the mean and dividing by the standard deviation, with these parameters derived only from the training set.

\textbf{Segmentation:} For consistency, we use a window size of 2 seconds (512 samples) for both fixed-length and CTXSEG segmentation, being the median of common window sizes used in seizure detection tasks~\cite{Xi_2022_TwoStage,He_2022_Spatial,Thuwajit_2022_EEGWaveNet,srinivasan_detection_2023,qiu_lightseizurenet_2023,Zhu_2023_Automated,Song_2023_EEG,ingolfsson_minimizing_2024,awais_graphical_2024,li_end--end_2024,busia_reducing_2024,wang_epileptic_2024}. For fixed size segmentation, we investigate overlaps of 25\%, 50\% and 75\%. When we use CTXSEG segmentation (i.e. \VarFirst, \VarRand), we investigate significance level (alpha) of 0.001, 0.01 and 0.05, which represent common p-value significance levels for statistical testing using the paired t-test in the medical domain~\cite{Chu_1999_introduction,Cesana_2018_What,Ioannidis_2018_Proposal}. When we use fixed-length segmentation (i.e., Fixed), we use 50\% overlap representing the most robust fixed-length configuration found in our initial evaluation.

\textbf{Multi-channel segment boundaries:} Since CTXSEG discovers segment boundaries on a per-channel basis and the data contain 19 channels, we propose a heuristic for aggregating segment boundaries across channels based on majority voting. For this dataset, we consider a multi-channel segment boundary, a boundary that spans all 19 channels for a given time, when a boundary is discovered on at least 2 channels within a small time tolerance (i.e., 2 samples). We choose 2 channels on the basis that several subjects in the dataset contains focal seizures that are annotated for 2 channels. In addition, we enforce a minimum segment size equal to the window size for consistency with single-channel CTXSEG.

\textbf{Training method:} We use identical training parameters for all experiments and only vary the segmentation method used for preprocessing. This involves a learning rate of $2e^{-5}$ with the AdamW optimiser. Training takes place for a minimum of 10 epochs to a maximum of 1000 epochs, with early termination if the dev set loss fails to improve for 50 consecutive epochs. Evaluation is conducted using the checkpoint based on the lowest dev set loss.

\section{Segmentation used for seizure detection tasks}

\begin{table}[ht]
    \caption{Segmentation methods in recent seizure detection tasks}
    \centering
    \scalebox{0.9}{
    \begin{tabular}{c|c|c}
        \toprule
         \tcellbf{Reference}&  \tcellbf{Year}& \tcellbf{Segmentation\\method(s)}\\

         \midrule
         \makecell{\cite{Xi_2022_TwoStage}}&  
         \makecell{2022}& 
         \makecell{
            Fixed: 1 s (sliding-window).
         }\\
         
         \midrule
         \makecell{\cite{He_2022_Spatial}}&  
         \makecell{2022}& 
         \makecell{
            Fixed: 1 s (50\% overlap).
         }\\

         \midrule
         \makecell{\cite{Thuwajit_2022_EEGWaveNet}}&  
         \makecell{2022}& 
         \makecell{
            Fixed: 4 s (1 s overlap).
         }\\
         
         \midrule
         \makecell{\cite{srinivasan_detection_2023}}&  
         \makecell{2023}& 
         \makecell{
            Fixed: 1, 2, 4 s (non-overlapping).
         }\\

         \midrule
         \makecell{\cite{qiu_lightseizurenet_2023}}&  
         \makecell{2023}&  
         \makecell{
            Fixed: 2 s (overlapping with 1 s stride).
         }\\
         
         \midrule
         \makecell{\cite{Zhu_2023_Automated}}&  
         \makecell{2023}&  
         \makecell{
            Fixed: 4 s (70\% overlap).
         }\\
         
         \midrule
         \makecell{\cite{Song_2023_EEG}}&  
         \makecell{2023}&  
         \makecell{
            Fixed: 1 s (non-overlapping).
         }\\

         \midrule
         \makecell{\cite{ingolfsson_minimizing_2024}}&  
         \makecell{2024}&  
         \makecell{
            Fixed: 1, 4, 8 s (non-overlapping).
         }\\

         \midrule
         \makecell{\cite{awais_graphical_2024}}&  
         \makecell{2024}&
         \makecell{
            Fixed: 30 s (30\% overlap).
         }\\

         \midrule
         \makecell{\cite{li_end--end_2024}}&  
         \makecell{2024}&  
         \makecell{
            Fixed: 4 s\\
            (50\% overlap in train, non-overlap in test).
         }\\

         \midrule
         \makecell{\cite{busia_reducing_2024}}&  
         \makecell{2024}&  
         \makecell{
            Fixed: 8 s (1s overlap).
         }\\

         \midrule
         \makecell{\cite{wang_epileptic_2024}}&  
         \makecell{2024}&  
         \makecell{
            Fixed: 128, 256 samples\\
            (overlapping with stride of 32, 64 samples)\\
            at 173.61Hz.
         }\\
         \bottomrule
         \addlinespace[3pt]
        \multicolumn{3}{p{260pt}}{
            Summary of segmentation methods used in recently published seizure detection methods that use machine learning. Fixed-size segmentation is used in all cases.
        }
    \end{tabular}
    }
    \label{tab:recent_studies}
\end{table}

\newpage


\section*{References}
\bibliographystyle{IEEEtran}
\bibliography{refs}

\end{document}